\documentclass{article}

\usepackage{microtype}
\usepackage{graphicx}
\usepackage{booktabs} 
\newcommand{\assessmentstep}[2]{%
  \ifx\relax#1\relax\else
    \noindent\textbf{#1:} #2\par\medskip
  \fi
}

\usepackage{hyperref}

\renewenvironment{quote}
  {\list{}{\leftmargin=20pt\rightmargin=20pt}%
   \item\relax}
  {\endlist}

\usepackage[accepted]{icml2025}

\usepackage{amsmath}
\usepackage{amssymb}
\usepackage{mathtools}
\usepackage{amsthm}
\usepackage{subcaption}
\usepackage[capitalize,noabbrev]{cleveref}
\usepackage{multirow}
\usepackage{enumitem}
\renewenvironment{quote}
  {\list{}{\leftmargin=0.2cm \rightmargin=0.2cm}%
   \item\relax}
  {\endlist}
  
\usepackage{hyperref}
\icmltitlerunning{Can LLMs Reason in a Legally Meaningful Manner on ECtHR cases?}

\begin{document}

\twocolumn[
\icmltitle{Can LLMs Reason in a Legally Meaningful Manner? \\A Small-scale Study on European Court of Human Rights Cases}



\begin{icmlauthorlist}
\icmlauthor{Amogh Raina}{cs}
\icmlauthor{Ilias Chalkidis}{cs}
\icmlauthor{Daniel Hershcovich}{cs}
\icmlauthor{Henrik Palmer Olsen}{law}
\end{icmlauthorlist}

\icmlaffiliation{cs}{Department of Computer Science, University of Copenhagen, Denmark}
\icmlaffiliation{law}{Faculty of Law, University of Copenhagen, Denmark}

\icmlcorrespondingauthor{Amogh Raina}{amogh.raina@di.ku.dk}

\icmlkeywords{Machine Learning, ICML}

\vskip 0.3in
]



\printAffiliationsAndNotice{} 

\begin{abstract}
Reasoning has become a standard technique and feature for contemporary LLMs; however, its application and quality in the context of demanding legal-oriented tasks, such as legal case forecasting, remain underexplored. We investigate how LLMs reason in the context of legal case forecasting, using legal cases from the European Court of Human Rights (ECtHR) as a testbed. We evaluate OpenAI GPT 5.4, a recent top-tier LLM, by exploring alternative prompting strategies that are more or less suggestive of what counts as legally meaningful reasoning in the context of ECtHR jurisprudence. We present our findings derived from assessing the model's responses with both human and LLM evaluation. We find that the examined model scores far from ideal in legal reasoning--the model produces structurally complete but substantively shallow analyses, and that LLM-as-a-Judge evaluators are internally consistent yet align only weakly with our trained annotators, i.e., reliable but not a valid substitute for human evaluation. Overall, the expert-curated prompt leads to more comprehensive reasoning, which does not result in more accurate predictions compared to the other examined settings. Based on our findings, we urge the community not to rely solely on automated LLM-based evaluation and to avoid using task accuracy as an appropriate proxy for reasoning quality. Code and data: \url{https://github.com/amogh-raina/ICML-EtCHR-Reasoning}.

\end{abstract}

\section{Introduction}

Large Language Models (LLMs) that employ reasoning-often referred to as Reasoning Language Models (RLMs)-have demonstrated
strong performance across a wide range of complex instruction-following and problem-solving tasks, including mathematics and coding~\citep{openai2024openaio1card,guo2025deepseek}. 
LLM reasoning has been a critical research topic and so far has been substantially benchmarked in domains like logic, math, and code; nonetheless, its potential application to complex, legal-oriented tasks--such as legal case forecasting (judgment)--remains heavily understudied. 
In such tasks, reasoning is not merely a path to better predictive accuracy but a lens into the model's decision-making, which is an explainability factor beyond brute-force pattern matching.
Recent industry surveys indicate that a substantial proportion of legal professionals, estimated between 60\% and 70\%, now report using generative AI tools at least weekly for drafting, research, or document analysis. \citet{neelamegam2025recent} reported that LLMs seem to excel at legal text classification and summarisation tasks, but struggle with tasks requiring structured legal reasoning. 

Legal case forecasting (judgment) is inherently reliant on sophisticated reasoning that differs across fields of law, e.g., criminal law, contract/commercial law, and human rights law, and jurisdictions, e.g., national or supranational, such as ECtHR. Analyzing and interpreting case facts, connecting them to the model's internal legal knowledge or other input sources, and making predictive decisions all require an advanced reasoning capability.

\begin{figure*}
    \centering
    \includegraphics[width=0.7\linewidth]{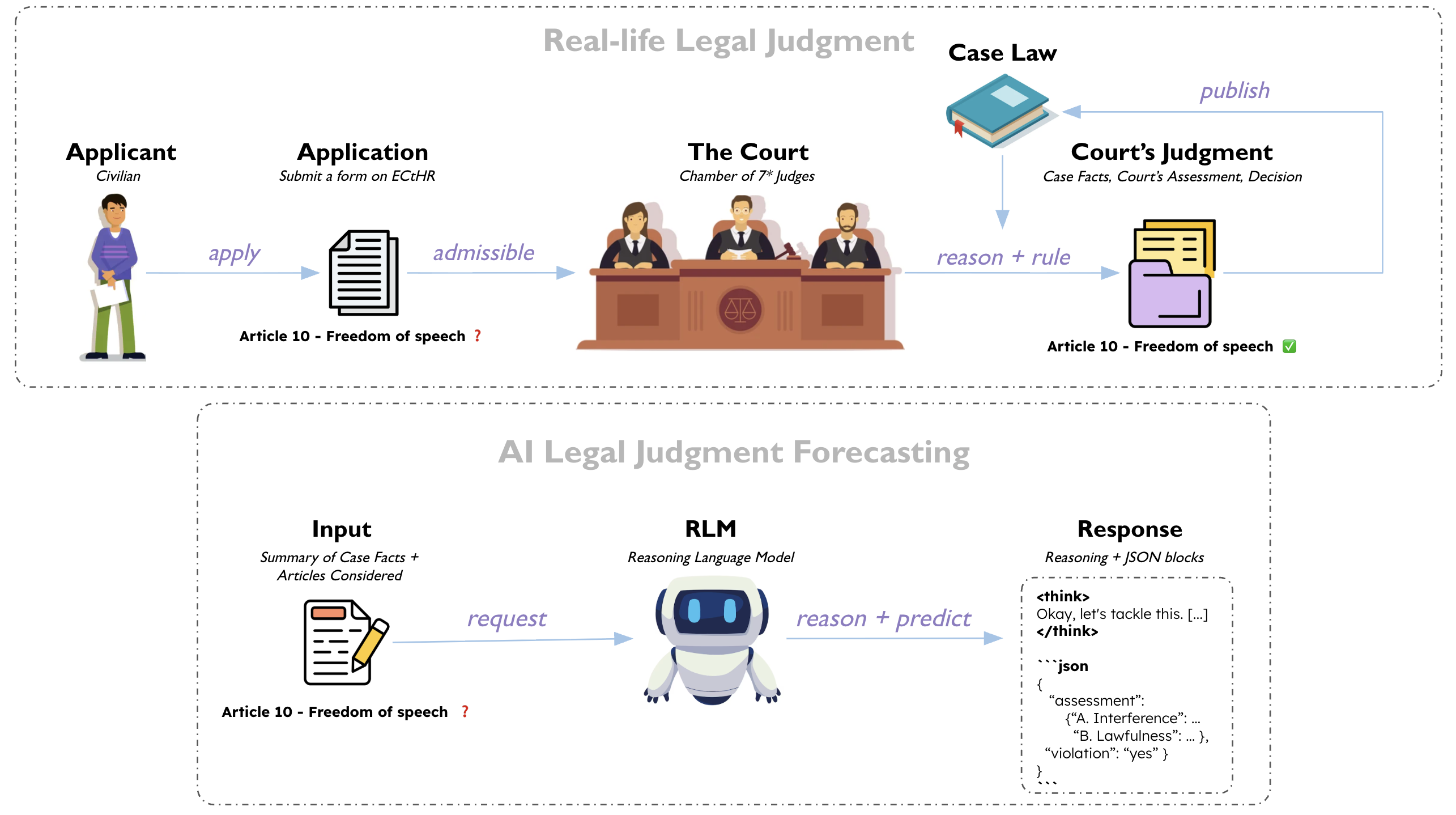}
    \vspace{-2mm}
    \caption{Cartoonish abstract depiction of the real-life ECtHR case workflow from case lodging to the court's judgment and its publishing; and of our LLM-driven legal judgment-forecasting experiment.}
    \vspace{-4mm}
    \label{fig:demo}
\end{figure*}

In this project, we investigate how LLMs reason in the context of legal case forecasting, using legal cases from the European Court of Human Rights (ECtHR) as a small-scale testbed: as in prior work~\citep{aletras2016predicting, chalkidis-etal-2019-neural,santosh-etal-2022}, the LLM assesses the facts and alleged violations and reasons about how the ECtHR would rule for or against the applicant. The general question we want to answer is: \textbf{\emph{Can LLMs reason in a legally meaningful manner?}} 

We approach this general question by addressing the following research questions:
\begin{enumerate}[label=RQ\arabic*:, leftmargin=*,topsep=0pt,itemsep=0pt]
    \item How competent is a recent top-tier LLM, such as OpenAI  GPT 5.4, in legal reasoning on ECtHR cases?
    \item How do different prompting strategies, i.e., the inclusion (or not) of relevant context, affect the model's reasoning performance?
    \item What is the quality of reasoning as assessed by human annotators and models (LLM-as-a-Judge)? And how does it differ among them?
\end{enumerate}

We focus on a short, targeted study of ECtHR cases related to Article 10 (Freedom of expression) of the European Convention of Human Rights (ECHR), which reads as follows:
\vspace{-3mm}
\begin{quote}
\vspace{-3mm}
\emph{
1. Everyone has the right to freedom of expression. This right shall include freedom to hold opinions and to receive and impart information and ideas without interference by public authority
and regardless of frontiers. This Article shall not prevent States from requiring the licensing of broadcasting, television or cinema enterprises.}

\emph{2. The exercise of these freedoms, since it carries with it duties and responsibilities, may be subject to such formalities, conditions, restrictions or penalties as are prescribed by law and
are necessary in a democratic society, in the interests of national security, territorial integrity or public safety, for the prevention of disorder or crime, for the protection of health or morals, for
the protection of the reputation or rights of others, for preventing the disclosure of information received in confidence, or for maintaining the authority and impartiality of the judiciary.}
\vspace{-3mm}
\end{quote}
\vspace{-3mm}
\paragraph{Contributions}

(a) We release a small dataset comprising 30 recent European Court of Human Rights (ECtHR) cases meant to assess LLMs' capabilities in legal judgment forecasting and reasoning (Section~\ref{sec:task_dataset}). \footnote{Code, prompts, model responses, LLM-judge evaluations and human annotations are available at \url{https://github.com/amogh-raina/ICML-EtCHR-Reasoning}.}
All cases have been officially published in the HUDOC database from April 2025 onwards.\footnote{We use a small dataset of recent cases to reduce (though not fully eliminate) overlap with the examined model's training data; in early experiments, we found that models memorize cases, especially popular ones that have been heavily discussed. As the model's knowledge cut-off (August~31, 2025) postdates part of our corpus, we discuss the residual contamination risk in the Limitations section.}

(b) We explore how the most recent OpenAI GPT 5.4 model (Knowledge cut-off: Aug 31, 2025) performs across different prompting settings: (a) \emph{out-of-the-box} with minor task guidance, or (b) under different settings where the prompt includes the ideal reasoning strategy step-by-step description \emph{explicitly} (verbatim), or \emph{implicitly}, providing the official multi-page guide related to the applicable article. 

(c) We evaluate the model's generated responses with a human evaluation (by law students) and models (LLM-as-a-Judge), focusing on the quality of the model's assessment (reasoning), rather than the model's decision (prediction). We also release, as part of our dataset (point a), the results of the human and LLM-based evaluations for future exploration or comparison with other LLM-as-a-Judge schemes.

\section{Examined Task and Data}
\label{sec:task_dataset}

The ECtHR hears allegations that Council of Europe member states (the defendant state(s)) have breached the European Convention of Human Rights (ECHR). Applicants lodge an application; if admissible, the court examines the case and rules on a decision, which is published and serves as precedent (Figure~\ref{fig:demo}).

We experiment with 30 ECtHR cases from the HUDOC database, related to Article 10 (Freedom of expression). For each, the dataset provides the \emph{facts} as presented in the decision (the factual events and relevant legal framework); Article 10 was \emph{allegedly violated} by the defendant state(s), and the court decided whether it was indeed \emph{violated} or not.

As in \citet{santosh-etal-2022}, the model receives the case facts and must reason and predict the (non-)violation of the article.\footnote{This is different from prior work~\citep{chalkidis-etal-2021-paragraph}, since the model is ``aware'' of the ECHR article under discussion, as is the court ruling the cases in real life.}  It generates both an \emph{assessment} (reasoning) and a \emph{decision} (Section~\ref{sec:experiment_settings}).
\vspace{-2mm}

\paragraph{Dataset Processing} To facilitate our experiments, we performed several pre-processing steps to extract the necessary information from the officially published court's judgment. 

First, we extract the \emph{facts of the case}, as represented in the court's decision, alongside the presented relevant legal framework, which summarizes the state of national law--of the defendant states(s)--related to the considerations of the examined case; this is used as the factual basis (input) provided to the model. The average length is 4,993 words.

Furthermore, we extract the \emph{court's assessment}, the part where the court reasons on how the facts should be interpreted under the law, covering interference, lawfulness, legitimacy, and proportionality, which we treat as our reference of expert reasoning. We keep only the Article-10 merits (excluding admissibility and out-of-scope articles) and extract its \emph{case-law references} (e.g., ``\emph{(See Axel Springer AG v. Germany [GC], no. 39954/08)}'') and \emph{factual references} (e.g., ``\emph{(See paragraphs 12-14 above)}''). On average, the assessment includes 25 factual and 10--15 case-law references.

\begin{figure*}
    \centering
    \includegraphics[width=1\linewidth]{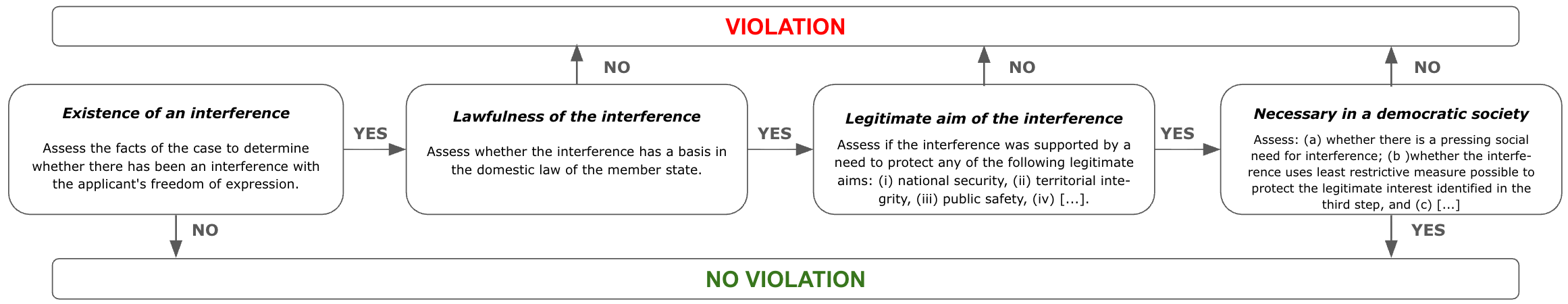}
       \caption{Appropriate reasoning strategy to assess ECHR cases related to Article 10 (Freedom of Expression).}
    \label{fig:strategy}
\end{figure*}

\subsection{Task Simplifications}
\label{sec:simplifications}

 The formulation of the task in our experiment does not replicate the court's flow in detail by any means. We follow a thin experimental setup, where, given curated facts, the models have to reason and predict the potential violations of the examined ECHR article. Here are the main differences:

\textbf{Input}: The input is the curated, summarised facts as written in the court's decision, i.e., the facts as the Court sees them after hearing both parties. The court itself has access to broader, unfiltered case files (factual background, parties' arguments, evidence); see~\citet{medvedeva2021automatic} for how facts are established.

\textbf{Background}: The models do not receive ECHR case law (precedent) verbatim. Rather than retrieve it via a RAG system---which risks cascading failures into the assessment---we provide the list of case-law references from the court's assessment as an oracle and ask the model to place them appropriately to support its arguments.

\textbf{Output}: In our experiments, the model produces the legal reasoning relevant to, and a prediction of the potential violation of the examined ECHR article (Article 10). In contrast, the ECtHR's ruling is more granular. The court considers the admissibility of a case as an initial pre-trial process and rules on potential damages, costs, and expenses that the defendant state must cover as reparations when the applicant (plaintiff) wins the case. Our model output does not include any of these components.

Despite all the aforementioned limitations, our framework is still the closest to the real judicial process followed by the Court, compared to the literature~\citep{aletras2016predicting, chalkidis-etal-2019-neural,santosh-etal-2022}, since our input includes the relevant legal framework, while we also request the model's assessment (reasoning) as a crucial part of supporting the model's decision that we later evaluate.

\subsection{Reasoning Desiderata}
\label{sec:desiderata}

\textbf{Qualities of legal reasoning}: Appropriate legal reasoning in the context of our work is considered one that (a) is grounded on the case facts, i.e., the interpretation of the facts is part of the reasoning, and (b) is relevant to the articles considered, i.e., the reasoning evolves around the application of the examined article(s) to the case facts in a legal-meaningful matter, i.e., following ECtHR standards as set out in its jurisprudence.

\textbf{Primacy of legal reasoning}: Appropriate legal reasoning prevails over predictive accuracy. A wrong prediction relative to the court's decision can arise from genuine ambiguity, narrative complexity, lack of context (Section~\ref{sec:simplifications}), or subjectivity~\citep{xu-etal-2023}; appropriate reasoning, by contrast, is the legally meaningful ``virtue'' that legitimises juridical bodies in democratic societies.

\section{Experimental Methodology}
\label{sec:experiments_methodology}

\subsection{Examined Model}

We experiment with OpenAI's GPT-5.4 model~\cite{gpt54}, the second-most capable LLM in the GPT series at the time, with a knowledge cutoff of August 31, 2025.\footnote{GPT-5.5, which we use as an evaluator (LLM-as-a-Judge), was released in the latter stage of our project.} We run the generator with \emph{medium} reasoning effort (exact per-setting values in Appendix~\ref{sec:models}). We focus on ChatGPT since it remains the most widely used model among legal professionals, ahead of other frontier LLM providers, making it the most representative choice for studying how legal practitioners actually encounter LLM-generated content.

\subsection{Prompting Settings}
\label{sec:experiment_settings}

We explore 3 alternative prompting settings:\footnote{In Appendix~\ref{sec:prompts}, we provide the instructions (prompts) for all settings verbatim.}

(a) \emph{Non-Curated}, where the model is instructed to assess and decide on a given case based on the facts and to format its output in a specific fashion, i.e., provide its assessment as a list of titled paragraphs alongside its decision (prediction) regarding the alleged violation (yes or no), but we do not provide any explicit information related to the appropriate reasoning strategy \textit{for the examined article}.

(b) \emph{Curated (Expert)}, where we provide a detailed step-by-step description of the appropriate reasoning strategy (Figure~\ref{fig:strategy}) related to the examined article--following ECtHR standards, curated by our legal expert. In this setting, the model is instructed to follow and apply the predefined reasoning strategy, and we expect a top-tier highly capable model to do this effectively.

(c) \emph{Curated (Guide)}, where we provide the official guide of the examined article (Article 10), which is part of the series of Case-Law Guides published by the European Court of Human Rights.\footnote{You can find the official guide \href{https://ks.echr.coe.int/documents/d/echr-ks/guide_art_10_eng}{here}.} This is a two-step process: (i) we first instruct the model explicitly to infer the appropriate reasoning strategy based on the guide's content; and (ii) we use the generated reasoning strategy as part of the follow-up prompt, where we request the model to assess and decide on a given case by following and applying the inferred strategy.

Exact API identifiers, version numbers, and reasoning-effort settings for the generator and all evaluators are reported in Appendix~\ref{sec:models}.

\subsection{Reasoning Evaluation}
\label{sec:evaluation}

We evaluate the formally disclosed case assessment (reasoning) in the model's output (evaluator models are run with \emph{high} reasoning effort). We ignore the ``hidden'' reasoning (or ``thinking'') that precedes the final answer, treating it as a scratchpad where the model explores the task, analogous to a student's notes during a test, or the Court's undisclosed deliberations.

\textbf{Human Evaluation}
As mentioned (Section~\ref{sec:desiderata}), our main goal is to assess the quality of the model's assessment (reasoning), rather than its decision (prediction). To do so, we recruit three senior law students, each with extensive training in ECtHR jurisprudence and working under an expert's supervision. For each case, we present to every annotator: (a) the original facts and the court's assessment as presented in the Court's judgment, and (b) the model's generated assessment (response) for two out of the three examined settings (a-c). The three annotators independently rate the same set of cases, which lets us report inter-annotator agreement and human--model alignment (Section~\ref{sec:reliability}).

\begin{table*}[]
    \centering
    \resizebox{\textwidth}{!}{
    \begin{tabular}{l|c|c|c|c|c|c|c|c|}
         \multirow{2}{*}{\textbf{Prompt Setting}} &  \multicolumn{2}{c|}{\textbf{Step Occurrence}} & \multicolumn{2}{c|}{\textbf{Step Comprehensiveness}} & \multicolumn{2}{c|}{\textbf{Overall Conciseness}} & \multirow{2}{*}{\textbf{Fact.\ Ref.\ F1}} & \multirow{2}{*}{\textbf{Accuracy}} \\
         & Human & Models  & Human & Models  & Human & Models & & \\
         \midrule
           A: Non-Curated & 0.96 $\pm$ 0.19 & 0.92 $\pm$ 0.28 & 3.95 $\pm$ 1.08 & 3.61 $\pm$ 1.44 & 3.87 $\pm$ 0.75 & 3.83 $\pm$ 0.94 & 0.51 $\pm$ 0.19 & \underline{0.82} \\
        B: Curated (Expert) & \underline{1.00 $\pm$ 0.06} & \underline{1.00 $\pm$ 0.00} & \underline{4.45 $\pm$ 0.79} & \underline{4.13 $\pm$ 0.82} & \underline{4.38 $\pm$ 0.67} & \underline{4.49 $\pm$ 0.54} & 0.51 $\pm$ 0.19 & 0.77 \\
        C: Curated (Guide) & 0.99 $\pm$ 0.11 & 0.99 $\pm$ 0.09 & 4.22 $\pm$ 0.88 & 4.00 $\pm$ 0.96 & 4.10 $\pm$ 0.71 & 3.92 $\pm$ 0.74 & \underline{0.53 $\pm$ 0.20} & 0.77 \\
        \midrule
        \emph{Majority baseline} (always ``violation'') & -- & -- & -- & -- & -- & -- & -- & 0.77 \\
    \end{tabular}
    }
    \caption{Results for each examined setting (A-C) across all metrics, averaged over all steps and the 22 distinct cases for GPT-5.4. For each setting, both Human (three annotators) and Models (LLM-as-a-Judge) ratings are reported. The always-``violation'' majority baseline attains $0.77$ accuracy; settings B and C match it exactly and A exceeds it by a single case, while reasoning-quality scores differ markedly across settings .i.e., reasoning quality does not track predictive accuracy (agreement with the court's decision; see Section~\ref{sec:reliability}).} 
    \vspace{-3mm}
    \label{tab:overall}
\end{table*}

Each annotator is asked to assess (annotate) the model responses based on the following criteria per model response:
\begin{enumerate}
    \item \emph{Step occurrence}: If each of the 4 steps of the ideal assessment (reasoning) strategy (Figure~\ref{fig:strategy}) occurs--is present--in the model's generated assessment. This is a binary label (yes or no). This criterion assesses whether the model follows a step, regardless of quality.
    \item \emph{Step comprehensiveness}: The comprehensiveness of each of the occurred steps in each response in a Likert-scale (1-5) compared to the court's assessment. If the expected step did not occur in the model's response, the score is 0. This criterion captures the overall quality of the reasoning step (Section~\ref{sec:desiderata}).
    \item \emph{Overall conciseness}: If the overall model's assessment for each response is concise, i.e., whether it contains no irrelevant information in a Likert-scale (1-5). This criterion captures a supplementary aspect of quality.
\end{enumerate}

This is a mixed evaluation setting, where each annotator assesses the quality of each model response given the ground truth (the court's assessment), but also implicitly ranks the two generated responses per step for the examined step's comprehensiveness, since both are presented in parallel. This allows us to consider individual--per setting--scores, but also pairwise comparisons, i.e., the ``win rate'' of one setting over the other.

\textbf{Sampling and comparability.} \textbf{Sampling and comparability.} Each case is shown as a pair of two of the three settings, fully balanced: each of the three pairs (A--B, A--C, B--C) is used for exactly 10 cases, so every setting is assessed on 20 cases and paired equally often with each other, with display order randomized. In total, the human evaluation covers 22 distinct cases as 30 (case, pair) instances (some cases are reused across pairs to keep the balance), all rated by the three annotators. This balance makes the per-setting averages in Table~\ref{tab:overall} directly comparable; the parallel presentation may still induce contrast effects on absolute scores, a further reason we emphasize rankings over absolute values (Section~\ref{sec:reliability}). 

\textbf{LLM-as-a-Judge Evaluation}
Similarly, we instruct three top-tier models (GPT-5.5, DeepSeek V4 Pro, and Claude Opus 4.7) to follow the same evaluation protocol as the human labeler. The models must respond with all relevant scores for each criterion, as instructed. The models assessed all three setting pairs across the dataset

\textbf{Other Automated Metrics}
We also compute the \emph{factual reference overlap} as a secondary metric of how well the model grounds its assessment in the case facts: we compare the factual paragraphs the model references (in the ``(see paragraph P)'' format) against those the Court references in its own assessment, using the F1-score to account for both recall and precision. Since the Court also cites paragraphs the model was never shown (e.g., from the party's submissions), we additionally restrict recall to paragraphs present in the facts as the reachable target set, and report \emph{groundedness}, the fraction of the model's references that point to a real facts paragraph.

For completeness, we also report \emph{predictive accuracy}, whether the model's prediction (violation or not) aligns with the court's ruling, though we do not treat it as a measure of reasoning quality (Section~\ref{sec:desiderata}).

\begin{figure*}[t]
    \centering
    \includegraphics[width=\linewidth]{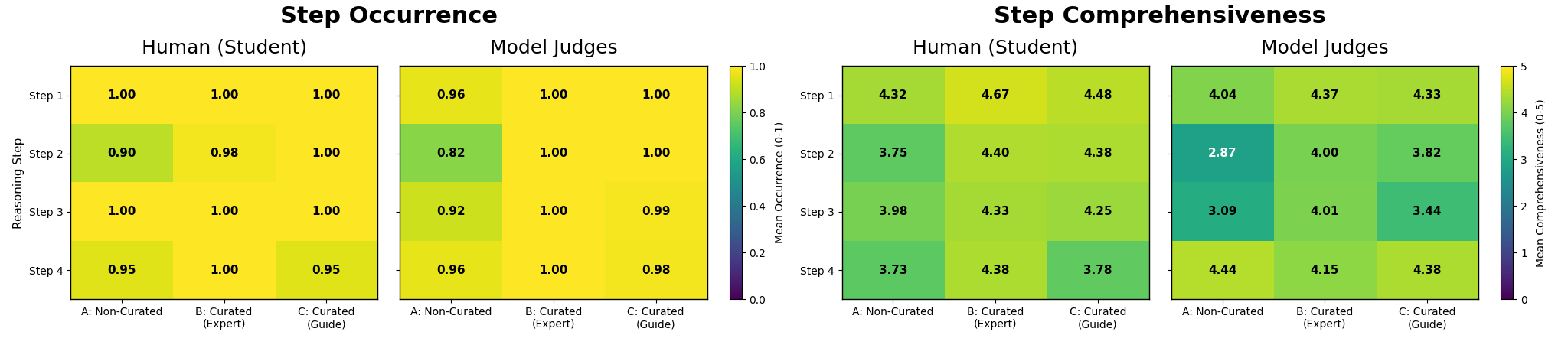}
    \vspace{-3mm}
    \caption{Results for step occurrence and comprehensiveness per step (1-4) across settings (A-C) .}
    \vspace{-3mm}
    \label{fig:heatmaps}
\end{figure*}

\section{Results \& Discussion}

\subsection{Hidden \& Output Reasoning}
\label{sec:hidden}

A reasoning model's response involves three distinct quantities: (i) \emph{internal reasoning}, a chain of computation the model is billed for; (ii) an emitted \emph{reasoning summary}, a short, provider-generated summary the API may return (for the OpenAI models we use, the raw chain is never exposed and this summary is often empty); and (iii) the \emph{output assessment}, the final answer. We evaluate only the output assessment (iii); by not assessing the model's ``hidden'' reasoning we mean the emitted summary (ii), which is at most a compressed, possibly unfaithful digest of the internal reasoning, so there is nothing reliable to score. This is a disclosure limitation, not a claim that the model reasons little

Indeed, the model reasons a great deal internally. Table~\ref{tab:placeholder} shows it spends $1.7$k--$3.5$k \emph{billed} tokens on internal reasoning (i), comparable to or larger than the visible output ($1.0$k--$1.3$k), yet almost none surfaces as text, the emitted summary is empty in $50/20/7\%$ of cases (A, B, C) despite those calls being billed thousands of reasoning tokens. Setting C triggers the most internal reasoning (partly its higher reasoning-effort; Appendix~\ref{sec:models}), while B is the most economical overall (shortest output, fewest total tokens).

\begin{table}[h]
    \centering
\resizebox{\columnwidth}{!}{
    \begin{tabular}{l|c|c|c}
         \bf Prompt Setting            &   \bf Internal reasoning &  \bf Output & \bf Total \\
         \midrule
A: Non-Curated       & 1709 $\pm$ 577  & 1341 $\pm$ 138 & 3050 \\
B: Curated (Expert)  & 2016 $\pm$ 644  & \underline{963 $\pm$ 109} & \underline{2979} \\
C: Curated (Guide)   & 3550 $\pm$ 1371 & 1230 $\pm$ 130 & 4780 \\
    \end{tabular}
    }
    \caption{Mean \emph{billed} tokens (from the API usage) for the model's \emph{internal reasoning} (stage (i): an internal chain of computation, not returned as text) and its disclosed \emph{output} assessment (stage (iii)), with the total (completion tokens), per setting over the 22 cases. The internal reasoning is billed but not observable in the output; the model emits at most a brief, often empty, reasoning summary (stage (ii)), which we do not evaluate.}
    \vspace{-4mm}
    \label{tab:placeholder}
\end{table}

\subsection{Overall Quantitative Assessment}

Table~\ref{tab:overall} reports the overall scores per metric, aggregated over all reasoning steps and cases.

\textbf{Step occurrence}
As we observe based on the human evaluation, we have very similar high scores (0.94-0.98) across all settings, which means that the examined model, GPT 5.4, is performing the relevant steps in most cases. The model evaluation offers similar results (0.92-1.00) with the same ranking and slightly inflated scores for settings B and C. Notably, even the Non-Curated setting (A), which contains no explicit description of the reasoning strategy, still elicits the four doctrinal steps in $\sim$90--100\% of cases. ``Non-Curated'' therefore means that we do not scaffold the reasoning in the prompt, not that the model is unaware of the four-step test: the structure appears frequently because it is standard Article~10 doctrine and thus already internalised by the model. The role of the curated prompts (B, C) is to make the model apply the test \emph{reliably} (removing the residual $8$--$18\%$ of missed steps under A), not to teach it from scratch. 

\paragraph{Step comprehensiveness} 
When it comes to comprehensiveness, both the annotators and the model judges identify more meaningful differences across settings. All settings score between 3 (average) and 4 (good) in this metric, which means that their reasoning is far from ideal (complete) compared to the court's assessment, but, nonetheless, leans towards positive. The best-performing setting is B (expert-curated), followed by setting C (model-inferred guide-based), with setting A (non-curated) in third place, a ranking shared by the annotators and the judges. The two rater types agree on this ranking but not on the absolute level: with three annotators, the human comprehensiveness scores are, if anything, slightly \emph{higher} than the judges' (e.g., $4.45$ vs.\ $4.13$ for setting~B), and the A$\to$B gap is comparable for both ($0.50$ for humans, $0.52$ for judges). Absolute comprehensiveness scores are therefore rater-dependent; we rely on the ranking, and quantify rater agreement in Section~\ref{sec:reliability}.

\textbf{Overall conciseness}  For the overall conciseness, we have a very similar picture with the same ranking across examined settings (B$>$C$>$A), and scores in the range 3 (average) to  4 (good). Here, however, the judges disagree among themselves more than on the other criteria (DeepSeek V4 Pro is lenient while Claude Opus 4.7 is strict; Section~\ref{sec:reliability}), so no single leniency direction holds across judges relative to the annotators.

\textbf{Factual Reference Overlap} The F1 between the model's and the Court's factual references is close to $0.51$--$0.53$ across settings, but this is driven by low \emph{precision}---the model references $\sim$21 paragraphs per case versus the Court's $\sim$13, i.e., it over-cites rather than by poor recall: when recall is restricted to paragraphs actually present in the facts, it is high ($0.81$--$0.86$), and \emph{groundedness} is $1.00$ (no hallucinated paragraph numbers across any response). The output-length instruction (``up to 1000 words'') is a soft guideline, not a hard token limit; since the model over-cites, its coverage is not suppressed by length. The model thus grounds its references reliably and recovers most facts the Court relied on, but hedges by over-citing.
\paragraph{Predictive Accuracy} All settings reach $77$--$82\%$ accuracy, but these differences are not meaningful: the corpus is class-imbalanced ($77\%$ of cases find a violation), and a trivial ``always-violation'' baseline attains $0.77$-exactly the accuracy of settings B and C, with A exceeding it by a single case (pairwise \mbox{McNemar} tests are non-significant, $\leq\!1$ discordant case; bootstrap $95\%$ CIs overlap fully). Crucially, accuracy does not track reasoning quality: the best-reasoning setting (B) sits exactly on the baseline, and at the case level comprehensiveness and prediction correctness are essentially uncorrelated ($r=0.08$; Section~\ref{sec:reliability}). The accuracy the model achieves thus reflects the majority class rather than the substantive legal test, so accurate predictions without legally meaningful reasoning are of little value (Section~\ref{sec:desiderata}).settings where reasoning matters.

\subsection{Step-wise Quantitative Assessment}

In Figure~\ref{fig:heatmaps}, we present step-wise results for occurrence and comprehensiveness to better understand which steps are harder to perform. We observe that step 2, i.e., assessing the lawfulness of the interference, is the step most frequently missed: it occurs in only $\sim$90\% of setting-A cases according to the annotators (and $\sim$82\% according to the model judges), making it the most fundamental gap identified by both. Elsewhere, misses are only occasional---the annotators record step 2 missing in setting B ($0.98$) and step 4, i.e., assessing the necessity of the interference in a democratic society, missing in settings A and C ($0.95$) while step 1 and step 3 occur in essentially all cases across settings.

Moving to step comprehensiveness, we first observe that the comprehensiveness of step 2 for setting A is clearly affected (penalized) by the fact that the examined model did not perform this step in roughly $10$--$18\%$ of the cases (per the annotators and the judges, respectively); but poor comprehensiveness is not merely a matter of not performing the step always, but rather \emph{how} the step was performed, with some interesting variations between model and human evaluation. On the one hand, the model judges score the comprehensiveness of step 3 lower than the rest for settings A and B, signifying considerable deviation from the court's assessment, while the annotators identify more profound inconsistencies in step 4 across all settings. In other words, the model reasoning is poorer for the latter (harder to assess) steps (3 and 4), overall.

\begin{figure*}[t]
    \centering
    \includegraphics[width=\linewidth]{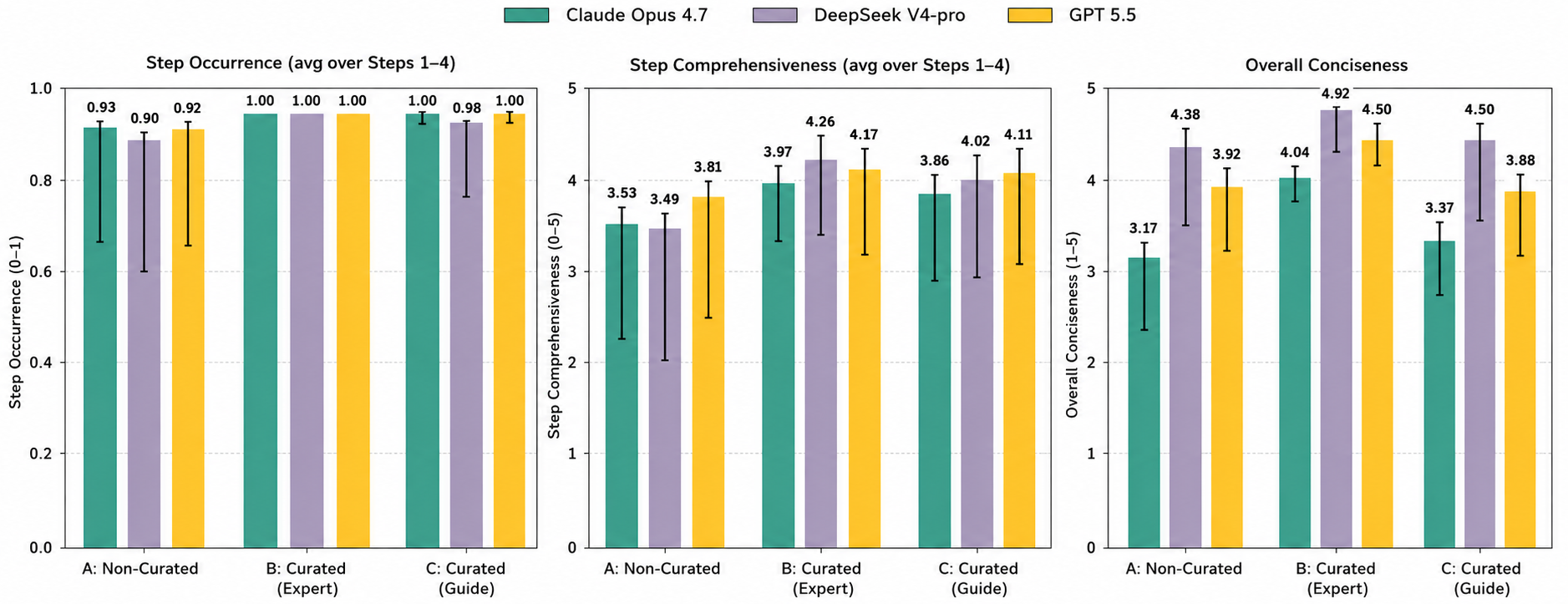}
    \vspace{-6mm}
    \caption{Results per Setting (A-C) across the three model judges (Claude Opus 4.7, DeepSeek V4 Pro, and GPT 5.5).}
    \vspace{-2mm}
    \label{fig:cros_judge}
\end{figure*}

\subsection{Cross-judge Comparison}

In Figure~\ref{fig:cros_judge}, we compare the three model judges. All judges agree on the ranking of settings (B$>$C$>$A) for every criterion, with only minor scoring differences on occurrence and comprehensiveness. They diverge most on overall conciseness, where DeepSeek V4 Pro is consistently the most lenient ($4.4$--$4.9$) and Claude Opus 4.7 the strictest ($3.2$--$4.0$). The \emph{ranking} across settings is thus robust across judges even though their absolute scores are not; we quantify how each judge aligns with the human annotators next (Section~\ref{sec:reliability}).

\subsection{Reliability and Validity of the Evaluation}
\label{sec:reliability}

Since our central claims rest on the human evaluation, we assess both its \emph{reliability} (do the three annotators agree?) and the \emph{validity} of the LLM judges as a proxy for human assessment (do the judges agree with humans?).

\begin{table}[t]
\centering
\resizebox{\columnwidth}{!}{
\begin{tabular}{l c c}
\multicolumn{3}{c}{\textbf{(a) Human inter-annotator agreement (3 annotators)}} \\
\midrule
Criterion & Chance-corr. & Agreement \\
\midrule
Step occurrence & $\kappa=0.14$ & $97\%$ (exact) \\
Step comprehensiveness & $\alpha=0.09$ & $87\%$ (within-1) \\
Overall conciseness & $\alpha=0.11$ & $91\%$ (within-1) \\
\bottomrule
\end{tabular}
}

\vspace{2mm}
\resizebox{\columnwidth}{!}{
\begin{tabular}{l c c c c c}
\multicolumn{6}{c}{\textbf{(b) Judge $\leftrightarrow$ human-consensus alignment}} \\
\midrule
& \multicolumn{2}{c}{Compr.} & \multicolumn{2}{c}{Concise} & Occ. \\
Judge & $\rho$ & bias & $\rho$ & bias & \%agr \\
\midrule
Claude Opus 4.7 & 0.16 & $-0.26$ & 0.24 & $-0.54$ & 97 \\
DeepSeek V4 Pro & 0.24 & $-0.05$ & 0.28 & $+0.44$ & 96 \\
GPT-5.5 & \underline{0.33} & $-0.01$ & 0.23 & $+0.01$ & 98 \\
\midrule
\emph{Judge--judge} $\alpha$ & \multicolumn{2}{c}{0.41} & \multicolumn{2}{c}{0.08} & \\
\bottomrule
\end{tabular}
}
\caption{Reliability and validity of the evaluation. \textbf{(a)}~Agreement among the three annotators: a chance-corrected coefficient (Fleiss' $\kappa$ for binary occurrence; Krippendorff's ordinal $\alpha$ for the $1$--$5$ criteria) alongside raw agreement (exact for occurrence, within-one-point for the $1$--$5$ criteria). The low chance-corrected values are a prevalence / restricted-range artifact---raw agreement is high (see text). \textbf{(b)}~Alignment of each LLM judge with the human consensus: Spearman $\rho$ and mean signed bias (judge $-$ human; positive $=$ more lenient) for comprehensiveness and conciseness, and \% agreement on occurrence.}
\label{tab:reliability}
\vspace{-3mm}
\end{table}

\paragraph{Do the annotators agree?} Almost always. The three annotators disagree on whether a step \emph{occurs} in only $11$ of $240$ cases ($97\%$ exact agreement), and on the $1$--$5$ criteria they land within one point of each other $87$--$91\%$ of the time (Table~\ref{tab:reliability}(a)). The chance-corrected coefficients ($\kappa$/$\alpha$) are nonetheless low ($0.09$--$0.14$). This is a known artifact: because almost every rating sits at the top of the scale (a step almost always occurs; quality clusters at $4$--$5$), that much agreement is already expected by chance, so little is left for $\kappa$/$\alpha$ to credit. Agreement is thus strong on the judgments themselves and on the \emph{ranking} of settings (all three annotators rank $B>C>A$), even if the exact score on the $1$--$5$ scale is somewhat rater-dependent---as, we now show, it is for the LLM judges too.

\paragraph{Can the LLM judges replace the humans? (reliability is not validity)} To check, we form a human \emph{consensus} score per case by averaging the three annotators, and compare each judge to it (Table~\ref{tab:reliability}(b)). The result is striking: the three judges agree with \emph{each other} far more than the humans agree among themselves (judge--judge $\alpha=0.41$ vs.\ human--human $\alpha=0.09$ on comprehensiveness), yet each judge correlates only weakly with the human consensus ($\rho=0.16$--$0.33$). The judges are consistent, then, but consistent with a \emph{shared model pattern} rather than with human judgment: \emph{reliability is not validity}. Among the three, GPT-5.5 tracks the human ranking best ($\rho=0.33$), Claude Opus 4.7 is the strictest (it scores below the humans), and DeepSeek V4 Pro the most lenient on conciseness; all match the human occurrence labels almost perfectly ($96$--$98\%$). This cautions against using an LLM judge as a drop-in substitute for human evaluation.

\paragraph{Does better reasoning mean a correct prediction?} No. Pooling all $60$ (case, setting) instances that received human scores, comprehensiveness and prediction correctness are essentially uncorrelated ($r=0.08$, $p=0.55$): correctly and incorrectly predicted cases have almost the same average comprehensiveness ($4.27$ vs.\ $4.20$). Reasoning quality thus carries essentially no signal about whether the model's prediction matches the court---our central decoupling claim, now measured across cases rather than over just three per-setting points.

\subsection{Qualitative Analysis}
\label{sec:qual}
Although step-comprehensiveness scores vary between 3 and 4 across raters, to understand the limitations of the model's reasoning we have to compare the model outputs against the actual court judgments.

One example is the assessment of lawfulness (step 2) in the case of Tergek v Türkiye (Case 1 in Appendix~\ref{sec:case-examples}), where the Court observes that ``\emph{it is not disputed between the parties that the interference was prescribed by law. Accordingly, it accepts that the interference complained of by the applicant had a legal basis under domestic law, namely either section 62 or section 68(3) of Law no. 5275.}''. 

The first examined setting (expert-curated) performs a thorough analysis of lawfulness, but only provides a vague conclusion: ``\emph{This strongly supports a finding that the interference was not lawful in the Convention sense.}'' The Court's conclusions are always firm: Either the interference was lawful, or it wasn't. If it was not lawful, the model should have predicted a violation for that reason, but it did not; instead it moved to the next steps. The score is low given the vagueness. Another issue is that the model places emphasis on the assessment performed during the national trial, where a lower Turkish court found the interference unlawful. This assessment, however, was overturned by the Turkish Constitutional Court. The fact that different court instances in Turkey assessed the case differently may have impacted the model's assessment, explaining the vagueness. Still, agreement og disagreement between national courts should not affect the reasoning; the Court must perform its own assessment and present a firm conclusion. Furthermore, the model's finding does not mirror the Court's finding, but this may be because the Court's argument for finding that the lawfulness requirement is fulfilled is based on information that was not available to the model, e.g., the parties' submissions (i.e. the information that the lawfulness of the interference was not disputed by the applicant). 
The second examined setting (non-curated) also gets the formal legal basis for the interference right, but gets a low score because it presents only a very short analysis and convolutes the lawfulness and legitimacy assessment (steps 2-3).

Both models score low (2) on step 4 (proportionality analysis). It is interesting to note here that the Court in pr. 63 identifies an important difference between printouts or photocopies on the one hand and officially published books or periodicals on the other. 
Specifically the court accept the Turkish Constitutional Court's assessment that printouts and photocopies sent to prisoners present risks to the security and order of the prison environment because of the possible infiltration of external communications within large number of printouts. 
Since it would be unreasonably burdensome on prison staff to review large amounts of photocopies or printouts for authenticity, the Turkish Constitutional Court and the ECtHR found that the interference was proportionate.


A reading of this difference is that the model fails to capture the tacit knowledge the Court applies here:
The Court, it could be said, understand the need for a workable system to ensure security in Turkish Prisons and therefore sees the interference as reasonable - The LLM's do not see this. 


\textbf{Recurring failure modes} Across the corpus, the model's errors recur in a few forms: vague or non-committal conclusions where the Court is firm; over-reliance on (sometimes overturned) domestic-court findings; shallow proportionality analysis that misses the Court's practical balancing; a majority-class bias toward predicting a violation; and over-citation of factual paragraphs. The first three concern the substantive legal judgment and are the most consequential.\vspace{-2mm}


\section{Limitations}

We identify a series of overall limitations in our work that future work should address:\vspace{-2mm}

\paragraph{Limited Scope} We consider ECtHR cases, and specifically, one of the many articles of the Convention (ECHR). Hence, our findings are limited to the restricted scope of our study. Nonetheless, our experiments are carefully curated to offer a meaningful, well-scoped evaluation. Future work shall explore a wider range of ECHR articles to better cover the evaluation of ECtHR reasoning.\vspace{-2mm}

\paragraph{Small Dataset} Our 30 Article-10 cases may not faithfully reflect the model's capability on the article broadly. We limit data contamination by using only very recent cases, but cannot fully rule it out: the model's cut-off (August~31, 2025) postdates part of our corpus (published from April~2025 onward), so roughly half of the cases could in principle fall within the training window. We expect memorisation to be limited, as these are recent, less-discussed judgments, unlike the popular older cases that motivated our recency constraint, but note this as a circular trade-off, since older, safely out-of-distribution cases require older, non-state-of-the-art models.\vspace{-2mm}

\paragraph{Single Model} We consider a single model, namely GPT 5.4. Hence, our study does not capture the capabilities of other LLMs to address the examined task. Nonetheless, we believe that using a top-tier, popular model is a good proxy for the state-of-the-art in the field.\vspace{-2mm}

\paragraph{LLM-Judge models} We note that the generator (GPT-5.4) and one of the three judges (GPT-5.5) share the same developer (OpenAI); to mitigate any resulting self-preference bias, we deliberately include two additional judges from different developers (Anthropic's Claude Opus 4.7 and DeepSeek V4 Pro) and report per-judge results (Figure~\ref{fig:cros_judge}), which expose single-judge idiosyncrasies rather than relying on any one evaluator.
\vspace{-2mm}

\paragraph{Human Annotators} We collect annotations from three senior law students, which lets us report inter-annotator agreement and human--model alignment (Section~\ref{sec:reliability}). They agree strongly on coarse judgments and on the ranking of settings, but the fine-grained $1$--$5$ scale is rater-dependent. All three are law students working under expert supervision rather than independent legal experts; the expert layer of our study is the supervision and the qualitative doctrinal analysis (Section~\ref{sec:qual}), while independent expert \emph{quantitative} annotation and a larger annotator pool remains valuable future work.\vspace{-2mm}

\section{Ethical and Societal Implications}

Our findings carry practical risks for deploying LLMs in legal settings. First, a model's \emph{disclosed} assessment need not reflect its actual internal computation: for the closed model we study, substantial hidden reasoning is billed but never returned (Section~\ref{sec:hidden}), so a fluent assessment gives no guarantee that its stated reasons produced the decision, a transparency concern for closed models in high-stakes use. Second, our reliability--validity gap (Section~\ref{sec:reliability}) cautions against automating the evaluation itself: LLM judges agree with one another far more than with trained annotators and over-rate the hardest (proportionality) step, so using them as the sole arbiter of ``good legal reasoning'' risks entrenching a shared model bias under a veneer of consensus. Third, the model's tendency to default to the majority outcome could, if deployed, systematically misjudge the minority of cases where restrictions on expression are in fact justified. Such systems should therefore support, not replace, human legal judgment, and automated evaluation should be validated against, not substituted for, expert assessment.\vspace{-2mm}

\section{Related Work}
\label{sec:related_work}

Building on a line of work on legal judgment prediction over ECtHR cases \citep{aletras2016predicting, chalkidis-etal-2019-neural, santosh-etal-2022}, recent work has stress-tested LLMs on free-text legal reasoning. \citet{shi-etal-2025-legalreasoner} proposes step-wise verification and correction over Hong Kong court cases, framing legal judgment prediction as a process-supervision problem with an expert-designed taxonomy of reasoning errors. \citet{chlapanis-etal-2025-greekbarbench} introduces a Greek Bar exam benchmark scored along three dimensions (Facts, Cited Articles, Analysis), paired with a meta-evaluation benchmark (GBB-JME) showing that LLM-judges align with human experts only when guided by span-based rubrics over expert-annotated ground-truths. \citet{juvekar-etal-2025-llms} report on Indian legal exams that frontier LLMs \emph{exceed} human toppers on objective multiple-choice papers but fail consistently on long-form reasoning, with recurring failure modes in authority, discipline, and forum-appropriate voice. \cite{enguehard-etal-2025-lemaj} improves reference-free LLM-as-Judge in legal Q\&A by decomposing answers into atomic Legal Data Points, reporting better correlation with human experts. 

Compared to this line of work, our study is deliberately narrow (one ECHR article, 30 recent cases, one model) but couples an expert-curated reasoning prompt with a parallel human/model evaluation that decouples reasoning quality from predictive accuracy; our step-wise analysis surfaces where that decoupling matters and cautions against task accuracy as a proxy for legal reasoning quality.

\section{Conclusion and Future Work}

We studied LLM legal reasoning on Article-10 ECtHR cases with GPT-5.4 under three prompting strategies (zero to expert-level guidance), evaluated by three annotators and three LLM judges. Our answer to the title question is \emph{not yet fully}: the model reliably reproduces the doctrinal structure, but its substantive reasoning remains shallow and the predictions it gets right largely track the majority outcome rather than sound legal analysis. The expert-curated prompt yields the most comprehensive reasoning yet no gain in accuracy, and the LLM judges prove reliable but only weakly aligned with human annotators. We thus urge the community not to rely solely on automated evaluation, nor to treat task accuracy as a proxy for reasoning quality.

In future work, we plan to extend the study to other ECHR articles (starting with Articles 3 and 11), enlarge the dataset, and evaluate several top-tier models, as well as recruit more (and more senior) annotators to further study inter-annotator agreement and expert reasoning quality.

\bibliography{example_paper}
\bibliographystyle{icml2025}

\appendix
\onecolumn
\section{Models and  cvz}
\label{sec:models}

All models were accessed through the OpenRouter API. Table~\ref{tab:models} reports the exact model identifiers, their role in our study, and the reasoning-effort setting used for each. GPT-5.4 serves as the generator; GPT-5.5, Claude Opus 4.7, and DeepSeek V4 Pro serve as the LLM-as-a-Judge evaluators. As these are very recently released models, some readers may be unfamiliar with them; we therefore provide their precise identifiers and versions here for reproducibility.

\begin{table}[h]
    \centering
    \resizebox{\textwidth}{!}{
    \begin{tabular}{l l l l l}
        \toprule
        \textbf{Role} & \textbf{Name (main text)} & \textbf{API identifier (OpenRouter)} & \textbf{Reasoning effort} & \textbf{Release date} \\
        \midrule
        Generator & GPT-5.4 & \texttt{openai/gpt-5.4} & medium & \emph{March 5, 2026} \\
        Evaluator & GPT-5.5 & \texttt{openai/gpt-5.5} & high & \emph{April 23, 2026} \\
        Evaluator & Claude Opus 4.7 & \texttt{anthropic/claude-opus-4.7} & high & \emph{April 16, 2026} \\
        Evaluator & DeepSeek V4 Pro & \texttt{deepseek/deepseek-v4-pro} & high & \emph{April 24, 2026} \\
        \bottomrule
    \end{tabular}
    }
    \caption{Exact model identifiers and settings. All models were accessed via the OpenRouter API. A, B, and C denote the Non-Curated, Curated (Expert), and Curated (Guide) prompting settings, respectively (Section~\ref{sec:experiment_settings}).}
    \label{tab:models}
\end{table}

\section{Instructions (Prompts)}
\label{sec:prompts}

We present the prompts (instructions) used across the different settings, as described in Section~\ref{sec:experiment_settings}:

\noindent\emph{A. Non-curated Instruction (Prompt)}

\begin{quote}
--------------------------------------------------------------------------------------------------------------------------------------

You are a legal assistant tasked with forecasting the outcome of cases brought to the European Court of Human Rights (ECtHR) by assessing the case facts.

\textbf{Task Instruction}

You are provided with a summary of the case facts related to a new case. In your analysis, you should focus only on information relevant to answering the question of whether there has been a violation of Article 10 of the European Convention of Human Rights (ECHR) regarding freedom of expression or not. You have to predict whether Article 10 of the ECHR has been violated in the examined case. To do so, you have to assess the facts of the case in light of ECtHR standards as set out in its jurisprudence. 

\textbf{Detailed Instructions}
\begin{itemize}
    \item Structure your case assessment as a list of titled paragraphs, separated step by step. It should be up to 1000 words.
    \item As part of your assessment, provide references to the factual paragraphs as part of the assessment, following the format used in ECHR proceedings, i.e.,  “(see paragraph P)”, where P is the number of the relevant paragraph as presented and numbered in the case facts, e.g., “(see paragraph 12)”.
    \item As part of your assessment, provide references to ECHR case law, following the format used in ECHR proceedings, i.e., “(see no. DDDDD/DD)”. You can only use the precedent cases from the list presented below.
    \item Do not address any other ECHR articles, except Article 10.
    \item Your response shall be in JSON format. The first key, named “assessment”, is your assessment (analysis), and the value shall be a list with titled paragraphs, one per reasoning step, i.e., {“Title of the paragraph”: “Content of the paragraph”}. The second one, named “violation”, is your prediction (violation [yes] or not [no])
\end{itemize}

\textbf{Relevant Case Law}

11882/10

2594/07

1172/12\\

\textbf{Examined Case Facts}

Case A V. D

1. [...]

2. [...]

3. [...]

--------------------------------------------------------------------------------------------------------------------------------------
\end{quote}

\noindent\emph{B. Curated (Step-by-step) Instruction (Prompt)}

\begin{quote}
--------------------------------------------------------------------------------------------------------------------------------------

You are a legal assistant tasked with forecasting the outcome of cases brought to the European Court of Human Rights (ECtHR) by assessing the case facts.

\textbf{Task Instruction}

[...]

\textbf{Assessment Strategy Instruction}

\begin{enumerate}
    \item “Existence of an interference”: Assess the facts of the case to determine whether there has been an interference with the applicant's freedom of expression. If you find no interference, predict that there has been no violation of the article and skip the rest of the steps; if you find that there has been an interference, proceed to step 2.
    \item “Lawfulness of the interference”: Assess if the interference is lawful. Assess whether the interference has a basis in the domestic law of the member state. If you find that the interference lacks a basis in the domestic law of the member state, predict that there has been a violation of the article and skip the rest of the steps; otherwise, proceed to step 3.
    \item “Legitimate aims of the interference”: Assess if the interference is legitimate. Assess if the interference was supported by a need to protect one of the following legitimate aims: (i) national security, (ii) territorial integrity, (iii) public safety, (iv) the prevention of disorder or crime, (v) the protection of health or morals, (vi) the protection of the reputation or rights of others, (vii) preventing the disclosure of information received in confidence, or (viii) maintaining the authority and impartiality of the judiciary. If you find that the interference does not support one of the legitimate aims, predict that there has been a violation of the article, and skip the next step; otherwise, proceed to step 4.
    \item “Necessary in a democratic society”: Assess if the interference is necessary in a democratic society. Assess: (a) whether there is a pressing social need for interference; (b) whether the interference uses the least restrictive measure possible to protect the legitimate interest identified in the third step, and (c) whether there are relevant and sufficient reasons to support the interference convincingly. If you find that the interference can be defended as necessary in a democratic society, then predict that there has been a violation of the article; if you find that the interference cannot be so defended, predict that there has been no violation.
\end{enumerate}

\textbf{Detailed Instructions}
\begin{itemize}
    \item Structure your case assessment as a list of 4 titled paragraphs, separated step by step. It should be up to 1000 words.
    \item As part of your assessment, provide references to the factual paragraphs as part of the assessment, following the format used in ECHR proceedings, i.e.,  “(see paragraph P)”, where P is the number of the relevant paragraph as presented and numbered in the case facts, e.g., “(see paragraph 12)”.
    \item As part of your assessment, provide references to ECHR case law, following the format used in ECHR proceedings, i.e., “(see no. DDDDD/DD)”. You can only use the precedent cases from the list presented below.
    \item Do not address any other ECHR articles, except Article 10.
    \item Your response shall be in JSON format. The first key, named “assessment”, is your assessment (analysis), and the value shall be a list with 4 titled paragraphs, one per reasoning step (1-4), i.e., {“Existence of an interference”: “...”}. The second one, named “violation”, is your prediction (violation [yes] or not [no])
\end{itemize}

\textbf{Relevant Case Law}

[...]\\

\textbf{Examined Case Facts}

[...]

--------------------------------------------------------------------------------------------------------------------------------------
\end{quote}

\noindent\emph{C. Curated (Guide) Instruction (Prompt)}

\noindent\emph{C1. Inferring Strategy from Guide Instruction}

\begin{quote}
--------------------------------------------------------------------------------------------------------------------------------------

You are a legal assistant tasked with assisting with tasks related to the European Court of Human Rights (ECtHR).

\textbf{Task Instruction}

You are provided with the official guide of Article 10 of the European Convention of Human Rights (ECHR), part of the series of Case-Law Guides published by the European Court of Human Rights. You have to extract from the presented guide (below) a step-by-step case assessment strategy (methodology) on how the Court’s judges shall assess the facts of each case in light of ECtHR standards as set out in the guide.

\textbf{Detailed Instructions}
\begin{itemize}
    \item Structure your case assessment strategy as a list of paragraphs, separated step by step. The steps shall describe sequential actions where the court has to assess the facts related to different factors. It should be up to 1000 words.
    \item Each step should have a title presenting the number of the step, e.g., ``1. Title of Step 1'', and the description of the step, i.e., what the judges have to assess in the specific steps and how they should proceed or not in further steps. For example, the description can be formulated like this: ``Assess the facts of the case to determine whether [...] If you find that [...] predict that there has been a violation of Article 10, and skip the next step; otherwise, proceed to the next step.''.
    \item Your response shall be in JSON format. The key, named “assessment\_strategy”, is your assessment strategy, and the value shall be a list with titled paragraphs, one per reasoning step (1-4), i.e., “1. Title of Step 1": “Description of Step 1”.
\end{itemize}

\texttt{Guide on Article 10 of the European Convention on Human Right}

[...]

--------------------------------------------------------------------------------------------------------------------------------------
\end{quote}

\noindent\emph{C2. Applying Inferred Strategy from Guide Instruction}

\begin{quote}
    --------------------------------------------------------------------------------------------------------------------------------------
    
\textbf{Task Instruction}

[...]

\textbf{Assessment Strategy Instructions}

\begin{enumerate}
    \item ``Identify the complaint structure: interference or positive obligation'': Assess whether there was an interference with freedom of expression, or instead a failure by the State to protect its effective exercise. Interference can take many forms: criminal conviction, damages, publication ban, confiscation, refusal of access, surveillance, source-disclosure order, dismissal from employment, disciplinary action, blocking of websites, or other measures capable of chilling future expression. Do not rely only on domestic labels; examine the practical effect on expression. Where proceedings ended without sanction, ask whether they still operated as a warning or deterrent. If there was no direct interference, assess whether the State had a positive obligation to secure effective freedom of expression, especially in journalism, private-employment settings, access to venues or information, and protection from intimidation or disruption. If you find neither interference nor a breached positive obligation, predict no violation of Article 10 and stop. If you find one of them, define the context of the speech (press, politics, judiciary, whistleblowing, internet, reputation, national security, elections, health or morals, etc.) and proceed.
    \item “Test the Article 10 § 2 gateways: lawfulness and legitimate aim'': Assess whether the interference was “prescribed by law”: the legal basis must exist in domestic law and have sufficient quality, meaning accessibility, foreseeability and safeguards against arbitrariness or abuse. Examine whether the rule was clear enough for the person concerned, taking account of the context, the field regulated, and the person’s professional status. Vague or contradictory rules, ad hoc interpretations, overbroad discretion, or lack of safeguards against arbitrary enforcement point toward unlawfulness. Then verify whether the interference pursued one or more legitimate aims listed exhaustively in Article 10 § 2, such as protection of reputation or rights of others, national security, prevention of disorder or crime, health or morals, confidentiality, or the authority and impartiality of the judiciary. If either lawfulness or legitimate aim fails, predict a violation of Article 10 and skip the next step. Otherwise proceed to necessity and proportionality.
    \item ``Assess necessity in a democratic society through contextual proportionality and balancing'': Assess whether the interference answered a pressing social need and was proportionate to the legitimate aim pursued. Start by fixing the margin of appreciation: it is narrow for political speech, matters of public interest, elections, press watchdog activity, and debate on the functioning of justice; wider for commercial speech, some morality cases, and general media-regulatory choices. Then ask whether domestic courts gave relevant and sufficient reasons and applied Convention standards to the facts. Examine the core contextual factors that fit the case: contribution to a debate of public interest; status and role of the speaker (journalist, politician, judge, lawyer, NGO, whistleblower, academic, ordinary citizen); target of the statement (politician, public official, private person, company, judiciary); distinction between facts and value judgments; existence of a sufficient factual basis; good faith and responsible journalism; method of obtaining the information and its veracity; content, tone, form, medium, audience and likely impact; and the severity and nature of sanctions, including chilling effect, imprisonment, damages, injunctions, dismissal, blocking or surveillance. Where two Convention rights conflict, especially Article 10 with Articles 8 or 6 § 2, conduct a fair-balance analysis using the Court’s specific criteria; if domestic courts already carried out that balancing in conformity with Strasbourg criteria, depart only for strong reasons. In thematic cases, add the guide’s specialised modules: source protection requires an overriding public interest and strong procedural safeguards; confidential-information and whistleblowing cases require attention to public interest, authenticity, reporting channels, harm caused and retaliation; access-to-information cases require that the request be instrumental to expression, concern information of public interest, come from a qualifying watchdog-type actor or equivalent, and concern ready and available material; national-security, disorder, health, morals and internet cases require close analysis of context, risk, reach and audience. If the reasons are not relevant and sufficient, the measure was not the least restrictive means, or the sanction is excessive, predict a violation; otherwise predict no violation.
\end{enumerate}
 
\textbf{Detailed Instructions}

[...]

\textbf{Relevant Case Law}

[...]\\

\textbf{Examined Case Facts}

[...]

--------------------------------------------------------------------------------------------------------------------------------------

\end{quote}

\noindent\emph{LLM-as-a-Judge Evaluation (Prompt)}

\begin{quote}
--------------------------------------------------------------------------------------------------------------------------------------

You are a legal assistant tasked with evaluating law students’ assessments of cases brought to the European Court of Human Rights (ECtHR).

\textbf{Task Instruction}

You are provided with a summary of the case facts (see “Case Facts” below) and the court’s assessment (see “Court’s Assessment” below) for an ECtHR case related to Article 10 of the European Convention of Human Rights (ECHR). You are also provided with the assessment from two law students for the very same case (see “Student A Response” and “Student B Response” below). You have to evaluate the quality of the students’ assessments in light of ECtHR standards as set out in its jurisprudence. The students’ assessments should follow the following reasoning strategy, step by step:

[...]

The students had to provide their assessment as a sequence of titled paragraphs, not necessarily following the exact same structure (presented above).

\textbf{Evaluation Criteria}

\begin{itemize}
    \item “Step occurrence”: Report if each of the 4 steps of the ideal assessment (reasoning) strategy (presented above) occurs–is present–in the student’s response. It is totally fine if the student performed the step, even if the step is part of a paragraph including more steps.  This is a binary label (yes or no).
    \item “Step comprehensiveness”: Report the comprehensiveness of each of the occurred steps in each student’s response in a Likert-scale (1-5, where 5 is greater than 1) in light of how the step is performed by the court’s assessment. If the expected step did not occur in the model’s response, the score is 0. If for a step, you score comprehensiveness for student A as 1, and for student B as 4, this implies that student B’s step was more comprehensive.
    \item “Overall conciseness”: Report if the overall student’s assessment is concise, i.e., if there is no irrelevant (redundant) information in each student’s response, as a whole, in a Likert-scale (1-5, where 5 is greater than 1).
\end{itemize}

\textbf{General Instructions}

\begin{itemize}
    \item Your response shall be in JSON format. The first key, named “evaluation”, is your evaluation (assessment), which has a list of JSON records, one for each student (“student\_a”, “student\_b”).
    \item For each student, there is a list for the assessment of the 4 steps (“step\_1”, “step\_2", “step\_3”, “step\_4"), where each step has a subfield (“occurrence”) where the value is 0 (no) or 1 (yes), and a sub-field (“comprehensiveness”), where the value is 1-5 (int), as described above (Evaluation Criteria).
    \item For each student, there is an extra field “overall\_conciseness”, where the values are 1-5 (int), as described above (Evaluation Criteria).
\end{itemize}

\textbf{Output Example}

[...]

\textbf{Case Facts}

[...]

\textbf{Court’s Assessment}

[...]

\textbf{Student A Response}

[...]

\textbf{Student B Response}

[...]

\end{quote}

\newcommand{\factpara}[2]{%
  \par\noindent\hangindent=2em\textbf{#1.}\enspace #2}

\newcommand{\step}[2]{%
  \noindent\textbf{#1:} #2\par\smallskip}

\newcommand{\violation}[1]{%
  \noindent\textit{Predicted violation:} \textbf{#1}}

\newcommand{\refparas}[1]{%
  \smallskip\noindent\textit{Court-referenced fact paragraphs:} #1}

\newpage
\section{Case Examples}
\label{sec:case-examples}

We present \textbf{2} representative cases below.
For each case we include: (i)~the case facts as provided to the model;
(ii)~the Court's Merits assessment (ground truth); and (iii)~model
assessments under each of the three prompting strategies
(Section~\ref{sec:experiment_settings}) using GPT-5.4.
Prompts~D and~E show assessments by Mistral Medium~3.5 under
Prompts~A and~B as a supplementary model comparison not discussed
in the main paper.


\subsection*{Case~1: TERGEK v. TÜRKİYE%
  \textnormal{\small\ (Application no. 39631/20), 29 April 2025}}

\subsubsection*{Case Facts}

{\footnotesize
\factpara{2}{The applicant was born in 1989 and is currently serving a prison sentence in Kocaeli T-Type Prison. He was represented by Mr S. Altıntaş, a lawyer practising in Kocaeli.}
\factpara{3}{The Government were represented by their Agent at the time, Mr Hacı Ali Açıkgül, former Head of the Department of Human Rights of the Ministry of Justice of the Republic of Türkiye.}
\factpara{4}{At the time of the events giving rise to the present application, the applicant was detained following his conviction for membership of an armed terrorist organisation described by the Turkish authorities as the ``Fetullahist Terror Organisation/Parallel State Structure'' (Fetullahçı Terör Örgütü / Paralel Devlet Yapılanması, hereinafter referred to as ``the FETÖ/PDY'').}
\smallskip\noindent\textit{withholding of letters sent to the applicant and related proceedings}\par
\smallskip\noindent\textit{First letter and related proceedings}\par
\factpara{5}{On 22 October 2018 the prison administration's letter-reading committee reviewed a letter sent to the applicant by his sister and found its enclosures to be objectionable (sakıncalı). The letter, which contained thirty-one pages of documents printed from the internet, was subsequently referred to the prison's Disciplinary Board for further examination.}
\factpara{6}{On the same day, the Disciplinary Board, citing section 68(3) of the Law on the execution of sentences and preventive measures (``Law no. 5275'' -- see paragraph 19 below), decided to withhold the letter. That decision was based on the grounds that the letter's enclosures contained statements which could potentially pose a threat to prison security, that it was unclear who had published the information or for what purpose, and that the information included phrases which could facilitate communication within the FETÖ/PDY organisation.}
\factpara{7}{On 26 October 2018 the applicant lodged an objection with the Kocaeli enforcement judge against the Disciplinary Board's decision. In his objection the applicant explained that he had injured his ankle on 1 October 2018 and had required a cast for several weeks. He stated that some of the withheld documents had been sent to him by his sister, a physiotherapist, and had contained information on various physiotherapy exercises he could do to assist with the rehabilitation of his ankle. The applicant further explained that the remaining documents related to a distance-learning course in real-estate management, which he was taking, and that he needed them in order to prepare for examinations.}
\factpara{8}{On 19 August 2019 the enforcement judge upheld the applicant's objection, citing the relevant principles and case-law of both the Court and the Constitutional Court. The judge ruled that withholding the documents solely on the grounds that the information contained in them was unclear from a general inspection, without a specific assessment of their actual content, had been unlawful in the light of the freedom of communication and expression.}
\factpara{9}{On 1 October 2019 the applicant was notified of that decision. On 25 October 2019 he received the letter and its enclosures.}
\smallskip\noindent\textit{Second letter and related proceedings}\par
\factpara{10}{On 18 December 2018, the letter-reading committee deemed another letter sent by the applicant's wife to be objectionable. The letter, containing sixty-one pages of documents printed from the internet, a one-page handwritten note and four pictures, was forwarded to the Disciplinary Board for further examination.}
\factpara{11}{On the same day, the Disciplinary Board, citing section 68(3) of Law no. 5275 and referencing a prior decision issued by the Administration and Monitoring Board on 4 November 2016 concerning the potential risks associated with allowing internet printouts to be handed over to prisoners, decided to withhold the documents in question from the applicant. It authorised the remaining items -- the one-page handwritten note and the four pictures -- to be handed over to him. The decision did not in any way address the content of the withheld documents.}
\factpara{12}{On 24 December 2018 the applicant objected to that decision before the enforcement judge. In his objection, he argued that the documents in question contained information on various physiotherapy exercises and real-estate management and were essential for both his rehabilitation and his ongoing education.}
\factpara{13}{On 27 August 2018, citing section 62 of Law no. 5275 (see paragraph 17 below), the enforcement judge dismissed the applicant's objection. The judge noted that the internet printouts could not be classified as ``books'' or ``correspondence'' under the relevant legislation as a result of their unknown origin and their susceptibility to external interference. The judge also noted that prisoners were free to access publications through the prison library.}
\factpara{14}{On 17 September 2019 the Kocaeli Assize Court, ruling on an objection lodged by the applicant, endorsed the reasoning provided by the enforcement judge.}
\smallskip\noindent\textit{INDIVIDUAL Application LODGED by the applicant WITH THE CONSTITUTIONAL COURT}\par
\factpara{15}{On 30 October 2019 the applicant lodged an individual application with the Constitutional Court, arguing, inter alia, that his right to respect for correspondence had been breached as a result of (i) the delayed delivery of the first letter and its enclosures, and (ii) the seizure of the second letter and its enclosures by the prison administration.}
\factpara{16}{On 16 June 2020 the Constitutional Court, sitting as a panel of two judges, dismissed those complaints as manifestly ill-founded. It referred to its leading judgment in the case of Diyadin Akdemir (application no. 2015/9562, 4 April 2018). The reasoning for its decision reads as follows:}
\smallskip\noindent\textit{``Having reviewed the application within the scope of the Constitutional Court's authority to examine individual applications, and having considered the documents submitted, it has been concluded that there was no interference with the fundamental rights and freedoms set forth in the Constitution, or that any interference that may have occurred did not constitute a violation of those rights (see, to the same effect, Diyadin Akdemir, application no. 2015/9562, 4 April 2018).''}\par
\smallskip\noindent\textit{RELEVANT LEGAL FRAMEWORK AND PRACTICE}\par
\smallskip\noindent\textit{RELEVANT DOMESTIC LEGISLATION}\par
\factpara{17}{Section 62 of Law no. 5275, entitled ``The right to receive periodical and non-periodical publications'', as in force at the material time, read as follows, in so far as relevant to the present case:}
\begin{itemize}
    \item ``1. The convicted person shall have the right to receive [any] periodical or non-periodical publication on payment of the [retail] price, provided that [such publications are not] prohibited by a court.
    \item 2. Newspapers, books and printed publications of public institutions, universities, professional organisations with the status of public institutions, foundations benefiting from a tax exemption [by decision of] the President of the Republic and associations working in the public interest shall be handed over free of charge to convicted persons. The textbooks used by convicted persons undergoing studies or training shall not be subject to inspection.
    \item 3. No publication endangering the security of the institution or containing obscene articles, writings, photographs, or comments shall be given to the convicted person....''
\end{itemize}

\factpara{18}{Following an amendment made by Law no. 7242 of 14 April 2020, subsection 3 of section 62 is now worded as follows:}
\begin{itemize}
    \item ``3. No publication that disturbs or endangers discipline, order, or security within the institution, [hinders] the achievement of the purpose of [rehabilitation] of convicted persons, or contains obscene articles, writings, photographs, or comments, shall be given to the convicted person.''
\end{itemize}
\factpara{19}{Section 68 of Law no. 5275, entitled ``The right to receive and send letters, faxes, and telegrams'', as in force at the material time, provided as follows, in so far as relevant:}
\begin{itemize}
    \item ``1. With the exception of the restrictions set forth in this section, convicted prisoners shall have the right, at their own expense, to send and receive letters, faxes, and telegrams.
    \item 2. The letters, faxes and telegrams sent or received by convicted prisoners shall be monitored by the reading committee in those prisons which have such a body or, in those which do not, by the highest authority in the prison.
    \item 3. [If] letters, faxes, and telegrams [to convicted prisoners] pose a threat to order and security in the prison, single out serving officials as targets, permit communication between members of terrorist or ... criminal organisations, contain false or misleading information likely to cause panic in individuals or institutions or contain threats or insults, they shall not be forwarded to [the addressee]. Nor shall [such letters, faxes, and telegrams] written by convicted prisoners be dispatched....''
\end{itemize}

\factpara{20}{Under section 116(1) of Law no. 5275, the provisions of the above sections may be applied to remand prisoners in so far as those provisions are compatible with the detention status of the prisoners concerned.}
\factpara{21}{Regulation 91 of the Regulations of 20 March 2006 on the management of prisons and the execution of sentences and preventive measures (``the Regulations''), published in the Official Gazette of 6 April 2006, as in force at the material time, provided as follows:}
\begin{itemize}
    \item ``1. Convicted prisoners shall have the right to send and receive letters, faxes, and telegrams at their own expense.
    \item 2. The letters, faxes and telegrams sent or received by convicted prisoners shall be monitored by the reading committee in those prisons which have such a body or, in those which do not, by the highest authority in the prison.
    \item 3. [If] letters, faxes, and telegrams [to convicted prisoners] pose a threat to order and security in the prison, single out serving officials as targets, permit communication for organisational purposes between members of terrorist or ... criminal organisations, contain false or misleading information likely to cause panic in individuals or institutions or contain threats or insults, they shall not be forwarded to [the addressee]. Nor shall [such letters, faxes, and telegrams] written by convicted prisoners be dispatched....''
\end{itemize}
\factpara{22}{For other provisions of domestic law relevant to the present case, se Halit Kara v. Türkiye (no. 60846/19, §§ 16-18, 12 December 2023), Mehmt Çiftci v. Turkey (no. 53208/19, §§ 10-15, 16 November 2021) and Osman and Altay v. Türkiye (nos. 23782/20 and 40731/20, §§ 13-19, 18 July 2023).}
\smallskip\noindent\textit{RELEVANT case-law OF THE CONSTITUTIONAL COURT}\par
\factpara{23}{In its judgment in the case of Diyadin Akdemir (application no. 2015/9562, 4 April 2018), the Constitutional Court examined an individual application challenging the prison authorities' refusal to provide a prisoner with photocopied documents sent to him by post. The court unanimously declared the application inadmissible as being manifestly ill-founded. It reasoned that photocopied documents were not covered by section 62 of Law no. 5275, which refers specifically to ``periodicals and non-periodicals''. Therefore, applying the same inspection criteria to photocopied documents as to periodicals and non-periodicals would impose an unreasonable burden on the prison administrations and domestic courts. The Constitutional Court further noted that photocopied documents could potentially raise copyright issues. The relevant parts of the judgment read as follows:}
\smallskip\noindent\textit{``20.... Regarding publications that are not subject to any prohibition orders, the Constitutional Court has stated that any interference with freedom of expression without justification, or without meeting the criteria established by it (Halil Bayık [GK], App. No: 2014/20002, 30/11/2017, §§ 28-43), would constitute a violation of Article 26 of the Constitution.}\par
\factpara{21}{It is clear that the inspection required in accordance with the principles and criteria outlined above cannot apply to photocopied documents, which do not fall within the scope of `periodicals and non-periodicals' as referred to n section 62 of Law no. 2575. Expecting photocopied documents sent to convicted and remand prisoners to be subject to inspection in accordance with the above-mentioned provision, in the light of the principles and criteria approved by the Constitutional Court, would place an unreasonable burden on the prison administrations and the lower courts.}
\factpara{22}{However, it should not be overlooked that documents in the form of photocopied books, as in the case at hand, may raise copyright issues. In this context, it cannot be argued that the intervention in question -- refusing to provide photocopied books to the applicant, a convicted person, on the grounds that inspection was not feasible -- was unnecessary in a democratic society.''}
}

\subsubsection*{Court's Assessment (Article~10 --- Merits)}

{\footnotesize
53.  The Court notes that the case concerns the applicant's request to receive information, in the form of internet printouts, which his wife sent to him by post and which the prison authorities refused to deliver. In that connection, the Court reiterates that, in general, prisoners continue to enjoy all the fundamental rights and freedoms guaranteed by the Convention, with the exception of the right to liberty. Thus, they continue to enjoy the right to freedom of expression (see Yankov v. Bulgaria, no. 39084/97, §§ 126-45, ECHR 2003-XII, and Tapkan and Others v. Turkey, no. 66400/01, § 68, 20 September 2007), which includes the right to receive information or ideas (see Mesut Yurtsever and Others v. Turkey, nos. 14946/08 and 11 others, § 101, 20 January 2015; Mehmet Çiftci, cited above, § 32; and Osman and Altay, cited above, § 40).

54.  The Court considers that the refusal of the national authorities to hand the documents in question over to the applicant amounted to an interference with his right to receive information and ideas (see Mehmet Çiftci, § 33, and Osman and Altay, § 41, both cited above).

55.  The Court observes that it is not disputed between the parties that the interference was prescribed by law. Accordingly, it accepts that the interference complained of by the applicant had had a legal basis under domestic law, namely either section 62 or section 68(3) of Law no. 5275.

56.  The Court further notes that the interference pursued legitimate aims within the meaning of Article 10 § 2 of the Convention, namely the protection of national security, the prevention of disorder and the prevention of crime.

57.  As regards the necessity of the interference, the Court reiterates the principles deriving from its case-law on freedom of expression, which are summarised in, inter alia, Bédat v. Switzerland ([GC], no. 56925/08, 29 March 2016) and Kula v. Turkey (no. 20233/06, §§ 45-46, 19 June 2018).

58.  In order to determine whether the interference with the applicant's right to freedom of expression has been convincingly justified in the present case, the Court must assess, in line with its case-law, whether the reasons provided by the national authorities to justify the interference were ``relevant and sufficient'' and whether the measure taken was ``proportionate to the legitimate aim pursued''.

59.  As to the assessment of whether the reasons provided were ``relevant and sufficient'', the Court notes that its task in exercising its supervisory jurisdiction is not to take the place of the competent domestic courts, but rather to review under Article 10 the decisions they have taken pursuant to their margin of appreciation. The domestic courts, given their constant contact with the realities of the country, are often better placed than an international judge to determine whether a fair balance was struck at a given moment. If the balancing exercise undertaken by the national authorities was carried out in compliance with the criteria established by the Court's case-law, serious reasons are required for the Court to substitute its opinion for that of the domestic courts (see Haldimann and Others v. Switzerland, no. 21830/09, §§ 54 and 55, ECHR 2015, and Bédat, cited above, § 54).

60.  In determining the proportionality of a general measure, such as the one at issue in the present case -- namely, the withholding of printed documents from a prisoner solely on the basis of their format -- the Court further reiterates that the quality of the judicial review of the necessity of the measure at the national level is of particular importance, including with regard to the application of the relevant margin of appreciation (see Animal Defenders International v. United Kingdom [GC], no. 48876/08, § 108, ECHR 2013 (extracts)). The Court has already held that a general measure is a more practical means of achieving the legitimate aim pursued than a provision allowing for case-by-case examination, as the latter system is likely to lead to considerable uncertainty, litigation, costs, and delays, or to discrimination and arbitrariness. Nevertheless, the way in which a general measure has been applied to the facts of a given case helps to reveal its practical impact and is therefore relevant to the assessment of its proportionality, making it an important factor to take into account (ibid.). It follows that the more persuasive the general justifications put forward in support of the general measure, the less importance the Court attaches to the impact of that measure in the particular case before it (ibid., § 109).

61.  Turning to the present case, the Court notes at the outset that the Constitutional Court, in its judgment in the Diyadin Akdemir case, set out the criteria that prison authorities must consider when examining photocopied documents sent to prisoners (see paragraph 23 above). Those criteria were reiterated and elaborated upon in detail in the written submissions by the Government (see paragraphs 46-51 above).

62.  The Constitutional Court explicitly stated that section 62 of Law no. 5275 referred specifically to ``periodicals and non-periodicals'' and that photocopied documents were not covered by that section. It held that applying the same inspection criteria to photocopied documents as to periodicals and non-periodicals would impose an unreasonable burden on prison administrations and the domestic courts. The Government, in their written submissions, also pointed out the significant risk of intra-organisational communication, particularly on account of the large volume of incoming documents relating to prisoners convicted of terrorism-related crimes.

63.  The Court notes that, although the receipt of photocopied or printed documents in prison was not explicitly regulated by domestic law, the Constitutional Court carried out a thorough and detailed assessment of the matter in the Diyadin Akdemir case. In its assessment, that court balanced the right of prisoners to access information and ideas with the duties and workload of the prison authorities, as well as the serious risks associated with intra-organisational communication. The Court recognises that reviewing a large volume of printed or photocopied documents, in addition to the regular publications sent to prisoners could indeed overwhelm prison staff, impede their duties, and place an excessive burden on the judiciary, including the Constitutional Court. It also acknowledges the inherent differences between printouts or photocopies and officially published books or periodicals, which typically undergo thorough reviews and regulatory controls prior to release to ensure compliance with legal standards. By contrast, printouts and photocopies sent to prisoners lack such pre-publication scrutiny, thereby presenting specific risks to the security and order of the prison environment, including the heightened risk of infiltration of certain external communications within large number of printouts (compare with Osman and Altay, cited above, § 53, which concerned publications sent to the applicants via postal services in disregard of the legal procedures laid down in that regard). The Court further notes that similar considerations were reflected in the decisions of the prison authorities (see paragraphs 11, 13 and 14 above).

64.  The Court has already noted the various means available to prisoners for obtaining publications in accordance with the relevant domestic law, namely obtaining publications upon request, internal access to publications issued by official authorities and certain organisations, books kept in prison libraries and school textbooks, and the possibility of receiving books as gifts on specific occasions (see the full text of the relevant provisions in Osman and Altay, cited above, §§ 14-17), which were still valid at the time of the events in the present case.

65.  The Court therefore finds that, in the circumstances of the case, it cannot be considered that the applicant's right to freedom to receive information and ideas was disproportionately restricted by his inability to access the printed material sent to him by post.

66.  In the light of the above, the Court finds it reasonable for the national authorities to regulate the manner in which prisoners may obtain photocopied or printed documents, a matter which falls within their margin of appreciation, in order to ensure the efficient functioning of all prison services. The Court therefore sees no reason in the present case to substitute its opinion for that of the Constitutional Court, which carried out a detailed and carefully balanced assessment of the competing interests involved and did not exceed its margin of appreciation.

67.  Consequently, the Court concludes that the retention of the printed documents that had been sent to the applicant did not constitute a violation of Article 10 of the Convention.

\refparas{11, 13, 14, 23, 46, 47, 48, 49, 50, 51}
}

\subsubsection*{Model Assessments}

\bigskip
\noindent\emph{A.\ Non-Curated (GPT-5.4)}

\begin{quote}\small
\step{Applicability of Article 10 and existence of an interference}{The measures complained of concern the applicant's receipt of a letter from his sister and a letter from his wife, together with printed documents relating, according to him, to physiotherapy exercises and a distance-learning real-estate course (see paragraphs 5, 7, 10 and 12). Even in prison, a detainee retains the right to receive information and ideas, and restrictions on incoming written material fall within Article 10 (see nos. 39084/97, 20233/06 and 66400/01). The withholding of the first set of documents for about one year, and the complete refusal to hand over the second set, therefore amounted to an interference with the applicant's Article 10 rights (see paragraphs 6, 9 and 11-14).}

\step{Prescribed by law and legitimate aim}{The interference appears to have had a basis in domestic law, namely sections 62 and 68 of Law no. 5275 and the relevant prison regulations (see paragraphs 17, 19 and 21). It also pursued at least on its face legitimate aims recognised under Article 10 § 2, notably the prevention of disorder or crime and the protection of prison security, especially since the applicant had been convicted of membership of the FETÖ/PDY (see paragraphs 4, 6 and 11). The central issue is therefore whether the authorities showed that the restrictions were necessary in a democratic society.}

\step{The first letter: lack of an individualised assessment}{As regards the first letter, the Disciplinary Board relied on general assertions that the internet printouts might threaten prison security, that their source and purpose were unclear, and that they might facilitate organisational communication within FETÖ/PDY (see paragraph 6). However, the applicant gave a concrete and innocent explanation: part of the material concerned ankle rehabilitation after an injury and part concerned study materials needed for examinations (see paragraph 7). Importantly, the enforcement judge later accepted his objection and expressly held that the withholding had been unlawful because the authorities had relied only on a general inspection and had not conducted any specific assessment of the actual content of the documents (see paragraph 8). The material was ultimately delivered only on 25 October 2019, around a year after it had been withheld (see paragraph 9). In the Court's case-law, prison restrictions based on abstract risks rather than a concrete examination of the material do not satisfy the requirement of relevant and sufficient reasons (see nos. 48876/08, 56925/08 and 14946/08).}

\step{The second letter: blanket exclusion of internet printouts}{The position is even clearer regarding the second letter. The Disciplinary Board withheld sixty-one pages of internet printouts simply by referring to section 68(3) and to a 2016 administrative decision about general risks associated with internet printouts; it did not address the content of the withheld documents at all (see paragraph 11). The applicant again explained that the material concerned physiotherapy and real-estate management and was important for his rehabilitation and education (see paragraph 12). Nevertheless, the enforcement judge dismissed the objection on purely formal grounds, stating that internet printouts were neither 'books' nor 'correspondence' because of their unknown origin and susceptibility to interference, and suggesting that the prison library was available instead (see paragraph 13). The assize court endorsed that reasoning (see paragraph 14). Such reasoning points to a near-automatic prohibition on a category of material, not a balancing exercise based on the specific content, the applicant's personal needs, or any demonstrated security risk. Under the Court's standards, a blanket or formalistic refusal to allow access to written material in prison is difficult to reconcile with Article 10 (see nos. 20233/06, 21830/09 and 66400/01).}

\step{Insufficient proportionality and weak procedural safeguards}{The proportionality assessment was particularly deficient because the impugned materials were said to serve rehabilitative and educational purposes, which are ordinarily compatible with the aims of detention rather than contrary to them (see paragraphs 7 and 12). The domestic authorities did not identify any concrete phrase, passage or image that endangered security, nor did they explain why less restrictive options were unavailable, such as examining the documents individually or removing only problematic pages. The Constitutional Court then dismissed the applicant's complaints with a brief formula referring to its previous case-law, without engaging with the specific facts of the present case, including the enforcement judge's finding that the first withholding had been unlawful and the absence of any content-based analysis for the second letter (see paragraphs 8, 11, 13 and 16). This weak review further supports the conclusion that the interference was not convincingly justified under Article 10 (see nos. 48876/08 and 56925/08).}

\step{Overall conclusion under Article 10}{Taking the two incidents together, the domestic authorities relied on vague security concerns and categorical objections to internet printouts, but failed to carry out the individualised examination required by Article 10. The first set of documents was withheld for a prolonged period despite later being found to have been blocked unlawfully, and the second set was withheld on an essentially blanket basis without any analysis of its content (see paragraphs 8, 9, 11, 13 and 14). In line with the Court's prison-expression jurisprudence, the reasons given were not relevant and sufficient, and the interference cannot be regarded as necessary in a democratic society (see nos. 39084/97, 20233/06, 14946/08 and 66400/01).}

\violation{yes}
\end{quote}

\bigskip
\noindent\emph{B.\ Curated Step-by-step (GPT-5.4)}

\begin{quote}\small
\step{Existence of an interference}{There was an interference with the applicant's Article 10 rights. The prison authorities first withheld a letter from his sister, including thirty-one pages of internet printouts, on 22 October 2018, and the applicant did not receive them until 25 October 2019 after a successful objection (see paragraphs 5-9). They also withheld a second set of materials sent by his wife, consisting of sixty-one pages of internet printouts, while delivering only the handwritten note and photographs (see paragraphs 10-14). Article 10 protects not only the imparting but also the receipt of information and ideas, including by prisoners, although imprisonment allows wider restrictions than in ordinary situations (see nos. 66400/01 and 21830/09). The delayed delivery of the first materials and the complete refusal of the second clearly affected the applicant's ability to receive educational and medical information which he said related to physiotherapy and a distance-learning real-estate course (see paragraphs 7 and 12).}

\step{Lawfulness of the interference}{The interference had some formal basis in domestic law, since the authorities relied on section 68(3) of Law no. 5275 for both decisions and the enforcement judge later referred, for the second letter, to section 62 of the same Law (see paragraphs 6, 11, 13 and 17-19). However, the quality of the legal basis appears deficient in its application to the present case. For the first letter, the Disciplinary Board used very general formulae about prison security and possible organisational communication, without identifying any concrete passage or making a specific content-based assessment; the enforcement judge later held that approach unlawful (see paragraphs 6-8). For the second letter, the Board relied on a prior abstract decision about risks linked to internet printouts and again did not assess the actual content at all (see paragraph 11). The enforcement judge then shifted the reasoning, saying internet printouts were neither `books' nor `correspondence' because of their unknown origin (see paragraph 13). That combination of generic reasoning, inconsistent legal characterisation, and absence of individual examination points to an arbitrary and insufficiently foreseeable interference, contrary to the Court's approach to the `prescribed by law' requirement (see nos. 14946/08 and 56925/08). This strongly supports a finding that the interference was not lawful in the Convention sense.}

\step{Legitimate aims of the interference}{The Government could plausibly argue that the measures pursued the prevention of disorder or crime and the protection of prison security, especially given the applicant's conviction for membership of the FETÖ/PDY and the prison authorities' concern that the materials might facilitate communication within that organisation (see paragraphs 4, 6 and 11). Those are legitimate aims under Article 10 § 2. Still, the legitimacy of the aims does not remove the need for an individualised demonstration that these particular materials created such a risk. Here, the authorities never explained why documents said to concern ankle rehabilitation and course preparation genuinely endangered prison order or security (see paragraphs 7 and 12).}

\step{Necessary in a democratic society}{The interference does not appear necessary in a democratic society. The domestic authorities did not show a pressing social need supported by relevant and sufficient reasons. The first set of documents was eventually delivered in full after the enforcement judge found the original decision unlawful because it had rested only on a general inspection and not on the actual content (see paragraphs 8-9). That later delivery undermines the claim of any real security risk. As to the second set, the authorities adopted a blanket refusal of internet printouts, without examining whether any page actually contained prohibited material and without considering less restrictive options such as checking, redacting, or withholding only problematic passages (see paragraphs 10-14). The applicant gave concrete and facially legitimate reasons for needing the documents for physiotherapy and education (see paragraphs 7 and 12), and section 62 itself shows special regard for textbooks used in studies (see paragraph 17). The enforcement judge's reference to access through the prison library was abstract and unsupported by any indication that equivalent materials were available there (see paragraph 13). Finally, the Constitutional Court dismissed the complaint by a formulaic reference to its earlier case-law and did not perform a real balancing exercise (see paragraph 16). In line with the Court's case-law requiring individualised and proportionate reasons for restricting prisoners' access to information (see nos. 20233/06, 48876/08 and 21830/09), the measures were disproportionate.}

\violation{yes}
\end{quote}

\bigskip
\noindent\emph{C.\ Curated Guide (GPT-5.4)}

\begin{quote}\small
\step{1. Interference with freedom to receive information}{The impugned measures disclose a clear interference with the applicant's Article 10 rights. The prison authorities withheld enclosures sent by the applicant's sister and wife, consisting of internet printouts which, according to the applicant, concerned physiotherapy exercises for his injured ankle and study materials for a distance-learning real-estate course (see paragraphs 7 and 12). The first set of documents was delivered only about one year later, after successful judicial challenge (see paragraphs 8-9), and the second set was never delivered (see paragraphs 11-14). Article 10 protects not only imparting but also receiving information, and prisoners retain that protection, subject to justified prison restrictions. Measures preventing access to written material, even in the prison context, are therefore interferences requiring justification (see nos. 39084/97, 14946/08, 56925/08).}

\step{2. Legal basis and legitimate aim}{The authorities relied on sections 68 and 62 of Law no. 5275, which regulate prisoners' correspondence and access to publications (see paragraphs 17 and 19). The aims invoked or inferable from the decisions were prison order and security, and prevention of communication within a terrorist organisation (see paragraphs 6 and 11), which are in principle legitimate under Article 10 § 2. However, the lawfulness of the interference is doubtful. In respect of the first letter, the enforcement judge expressly found the withholding unlawful because the administration had relied on a general and abstract assessment rather than the actual content of the documents (see paragraph 8). In respect of the second letter, the authorities oscillated between the regimes for correspondence and publications and ultimately treated internet printouts as neither, largely because of their form and supposed unknown origin (see paragraphs 11 and 13). Such formalistic and inconsistent classification weakens foreseeability and safeguards against arbitrariness. Still, even assuming the interference was prescribed by law, the decisive issue is necessity and proportionality.}

\step{3. Necessity and proportionality: first letter}{As to the first letter, the Disciplinary Board stated that the enclosures might threaten prison security, that their publisher and purpose were unclear, and that they contained phrases that could facilitate communication within FETÖ/PDY (see paragraph 6). Yet no concrete passage or identifiable risk was cited. The domestic court later accepted that withholding the material merely because its content was unclear upon a general inspection, without a specific assessment of what the documents actually contained, was incompatible with freedom of expression and communication (see paragraph 8). That finding is highly significant. Moreover, the delayed delivery deprived the material of much of its practical value, since the applicant had said the documents were needed both for ankle rehabilitation and exam preparation (see paragraph 7). A delay of nearly a year in giving access to allegedly innocuous educational and medical material cannot be regarded as a proportionate response to vague security concerns. Strasbourg case-law requires relevant and sufficient reasons and an individualised assessment, not speculative assertions (see nos. 66400/01, 20233/06, 56925/08).}

\step{4. Necessity and proportionality: second letter}{The second withholding appears even harder to justify. The Disciplinary Board withheld sixty-one pages of internet printouts by relying on section 68(3) and on a prior general decision warning of the risks of internet printouts in prisons, while authorising the handwritten note and photographs (see paragraph 11). Crucially, it did not address the content of the withheld material at all (see paragraph 11). The enforcement judge then rejected the applicant's objection on the basis that the documents were neither `books' nor `correspondence', because of their unknown origin and susceptibility to interference, and added that prisoners could use the library (see paragraph 13). That reasoning is overly formalistic and close to a blanket exclusion of internet printouts as a category. Article 10 does not permit such automatic restrictions where the authorities do not show why the specific material creates a genuine security problem. The applicant had directly explained that the documents related to physiotherapy and his studies (see paragraph 12), and there is nothing in the facts suggesting violence, propaganda, coded messages, or any concrete institutional risk. The library argument is also weak, because no court established that equivalent material was actually available there or that it would meet the applicant's immediate medical and educational needs. Blanket or category-based restrictions on access to written material, without content analysis, are generally incompatible with the proportionality test (see nos. 14946/08, 48876/08, 21830/09).}

\step{5. Quality of domestic review}{The domestic review did not cure the interference. Although the enforcement judge remedied the first withholding, that occurred only after a lengthy period and without compensation for the delay's effects (see paragraphs 8-9). As to the second withholding, the enforcement judge and Assize Court essentially endorsed a categorical exclusion based on form rather than substance (see paragraphs 13-14). The Constitutional Court then dismissed the complaint in a formulaic manner, stating either that there had been no interference or that any interference had not amounted to a violation, with a bare reference to its earlier case-law (see paragraph 16). That reasoning did not engage with the applicant's concrete explanations, the actual content of the documents, or the balancing required by Article 10. Where domestic courts fail to apply Convention standards and give relevant and sufficient reasons, the Strasbourg Court is more likely to find a violation (see nos. 39084/97, 20233/06, 56925/08).}

\step{6. Overall conclusion}{Even allowing the State a relatively wider margin of appreciation in prison-security matters, especially concerning a prisoner convicted of membership of a terrorist organisation (see paragraph 4), the restrictions here appear disproportionate. The decisive features are the absence of a specific content-based assessment, the reliance on speculative or generic security concerns, the formalistic exclusion of internet printouts as a category, the substantial delay in delivery of the first material, and the insufficiency of the domestic courts' reasoning (see paragraphs 6-16). In light of Article 10 principles requiring careful, individualised justification for restrictions on receiving information, the Court would likely conclude that the applicant's freedom of expression was violated (see nos. 66400/01, 14946/08, 56925/08).}

\violation{yes}
\end{quote}

\bigskip
\noindent\emph{D.\ Non-Curated (Mistral Medium 3.5)  \textnormal{\small(supplementary; not in main paper)}}

\begin{quote}\small
\step{Establishment of interference}{The withholding of two letters and their enclosures addressed to the applicant by the prison authorities constitutes an interference with his right to receive information, which falls within the scope of Article 10 of the Convention (see paragraphs 5-14).}

\step{Prescribed by law and legitimate aim}{The interference was based on sections 68(3) and 62 of Law no. 5275 and pursued legitimate aims under Article 10(2), namely the prevention of disorder and crime in the prison context (see paragraphs 6, 11, 17-19).}

\step{Lack of specific content assessment -- first letter}{The Disciplinary Board's decision to withhold the first letter was based on general concerns about prison security and unclear publication details, without a specific assessment of the documents' content (see paragraph 6). The enforcement judge later ruled this decision unlawful precisely for this reason, noting that withholding based solely on general inspection without content analysis violated freedom of communication and expression (see paragraph 8). The documents contained physiotherapy and educational material, which are particularly protected forms of information (see paragraph 7).}

\step{Blanket restriction -- second letter}{The second letter's enclosures were withheld based on a blanket policy against internet printouts, as evidenced by the reference to a 2016 decision, without any examination of their specific content (see paragraph 11). The enforcement judge and Assize Court upheld this on formal grounds, ignoring the substance of the materials (see paragraphs 13-14). The Constitutional Court's dismissal relied on a formalistic distinction between document types, rather than a proportionality assessment (see paragraph 16).}

\step{Disproportionate interference under ECtHR standards}{The ECtHR has established that restrictions on freedom of expression must be necessary in a democratic society, requiring a pressing social need and proportionality (see no. 48876/08). Blanket restrictions without individual content assessment are generally incompatible with Article 10 (see no. 56925/08). Here, neither withholding decision demonstrated a concrete, content-based justification. The legitimate purposes of the materials---medical rehabilitation and education---further underscore the disproportionate nature of the restrictions (see paragraphs 7, 12).}

\step{Failure of domestic review to meet Convention standards}{While the domestic enforcement judge eventually rectified the withholding of the first letter, the initial interference lacked a lawful basis under national standards (see paragraph 8). For the second letter, all domestic instances upheld the restriction based on the formal classification of the documents as internet printouts, rather than their actual content. This approach fails to meet the ECtHR's requirement for a specific, proportional assessment under Article 10 (see no. 21830/09).}

\violation{yes}
\end{quote}

\bigskip
\noindent\emph{E.\ Curated Step-by-step (Mistral Medium 3.5)  \textnormal{\small(supplementary; not in main paper)}}

\begin{quote}\small
\step{Existence of an interference}{The withholding of the applicant's incoming letters and their enclosures by the prison authorities clearly constituted an interference with his right to receive information under Article 10 of the Convention. The applicant was prevented from accessing materials sent to him, which affected his freedom of expression (see paragraphs 5, 6, 10, 11).}

\step{Lawfulness of the interference}{The interference had a legal basis in domestic law. The withholding of the letters was carried out pursuant to section 68(3) of Law no. 5275, which explicitly permits the restriction of correspondence that poses a threat to prison order or security or facilitates communication between members of terrorist or criminal organisations (see paragraphs 6, 11, 19). Additionally, the enforcement judge's reasoning for the second letter cited section 62 of the same law (see paragraph 13).}

\step{Legitimate aims of the interference}{The interference pursued legitimate aims under Article 10(2) of the Convention. The Disciplinary Board justified the withholding of the first letter on the grounds that its enclosures could pose a threat to prison security and facilitate communication within the FETÖ/PDY organisation (see paragraph 6). For the second letter, the Board relied on prior guidance concerning the risks associated with internet printouts being handed over to prisoners (see paragraph 11). These reasons align with the legitimate aims of preventing disorder or crime and protecting national security.}

\step{Necessary in a democratic society}{The interference was not necessary in a democratic society. Regarding the first letter, the domestic enforcement judge ruled that withholding the documents solely on the grounds that their content was unclear from a general inspection, without a specific assessment of their actual content, was unlawful (see paragraph 8). For the second letter, the Disciplinary Board's decision did not address the content of the withheld documents at all, instead applying a blanket policy against internet printouts (see paragraph 11). The applicant demonstrated that the materials served legitimate purposes related to his medical rehabilitation and education (see paragraphs 7, 12). The ECtHR has consistently required that restrictions on prisoners' correspondence be based on a specific, individualized assessment of the content and be proportionate (see no. 48876/08). The lack of such an assessment, combined with the availability of less restrictive measures (e.g., content review or provision of materials through approved channels), renders the interference disproportionate and unjustified.}

\violation{yes}

\end{quote}
\end{document}